\pdfoutput=1

\documentclass[11pt]{article}

\usepackage[final]{acl}

\usepackage{times}
\usepackage{latexsym}

\usepackage[T1]{fontenc}

\usepackage[utf8]{inputenc}

\usepackage{microtype}

\usepackage{inconsolata}

\usepackage{graphicx}

\usepackage{arydshln}

\usepackage{microtype}
\usepackage{hyperref}
\usepackage{url}
\usepackage{booktabs}
\usepackage{xspace}
\definecolor{darkblue}{rgb}{0, 0, 0.5}
\hypersetup{colorlinks=true, citecolor=darkblue, linkcolor=darkblue, urlcolor=darkblue}

\usepackage{graphicx}
\usepackage{mfirstuc}
\usepackage{multirow}

\newcommand{\predictor}[1]{next token predictor\xspace}
\newcommand{\predictors}[1]{next token predictors\xspace}
\newcommand{\prediction}[1]{next token prediction\xspace}
\newcommand{\worldmodel}[1]{world model\xspace}
\newcommand{\worldmodels}[1]{world models\xspace}
\newcommand{\worldmodeling}[1]{world modeling\xspace}
\newcommand{\agent}[1]{agent model\xspace}
\newcommand{\agents}[1]{agent models\xspace}
\newcommand{\agenting}[1]{agent modeling\xspace}
\newcommand{\rlhf}[1]{RLHF\xspace}
\newcommand{\nervecenterspan}[1]{anchor span\xspace}
\newcommand{\nervecenterspans}[1]{anchor spans\xspace}

\usepackage{enumitem,amssymb}
\newlist{todolist}{itemize}{2}
\setlist[todolist]{label=$\square$}
\usepackage{pifont}
\title{Predicting vs. Acting: \\ A Trade-off Between World Modeling \& Agent Modeling}

\author{*Margaret Li$^{1,2}$ \hspace{6.0pt} *Weijia Shi$^{1,2}$ \hspace{6.0pt} Artidoro Pagnoni$^{1,2}$ \\
  \textbf{Peter West$^{1,3}$ \hspace{6.0pt} Ari Holtzman$^{2,4}$} \\
  $^1$University of Washington \hspace{6.0pt} $^2$Meta \\
  $^3$University of British Columbia \\
  $^4$University of Chicago\\ 
  \texttt{\{margsli,swj0419,artidoro\}@cs.washington.edu} \\ \hspace{6.0pt} \texttt{pwest@cs.ubc.ca} \hspace{6.0pt} \texttt{aholtzman@uchicago.edu}}

\begin{document}

\maketitle

\begin{abstract}

\renewcommand\thefootnote{*}\footnotetext{Authors contributed equally}

\renewcommand*{\thefootnote}{\arabic{footnote}}
\setcounter{footnote}{0}
RLHF-aligned LMs have shown unprecedented ability on both benchmarks and long-form text generation, yet they struggle with one foundational task: next-token prediction. As RLHF models become \agents{} aimed at interacting with humans, they seem to lose their \emph{world modeling}---the ability to predict what comes next in \emph{arbitrary} documents, which is the foundational training objective of the Base LMs that RLHF adapts. 

Besides empirically demonstrating this trade-off, we propose a potential explanation: to perform coherent long-form generation, RLHF models restrict randomness via implicit \textit{blueprints}. In particular, RLHF models concentrate probability on sets of \emph{anchor spans} that co-occur across multiple generations for the same prompt, serving as textual scaffolding but also limiting a model's ability to generate documents that do not include these spans. We study this trade-off on the most effective current agent models, those aligned with RLHF, while exploring why this may remain a fundamental trade-off between models that \emph{act} and those that \emph{predict}, even as alignment techniques improve.

\end{abstract}

\begin{figure*}
\centering \footnotesize

\begin {minipage}{\textwidth} \centering
\includegraphics[width=0.9\linewidth]{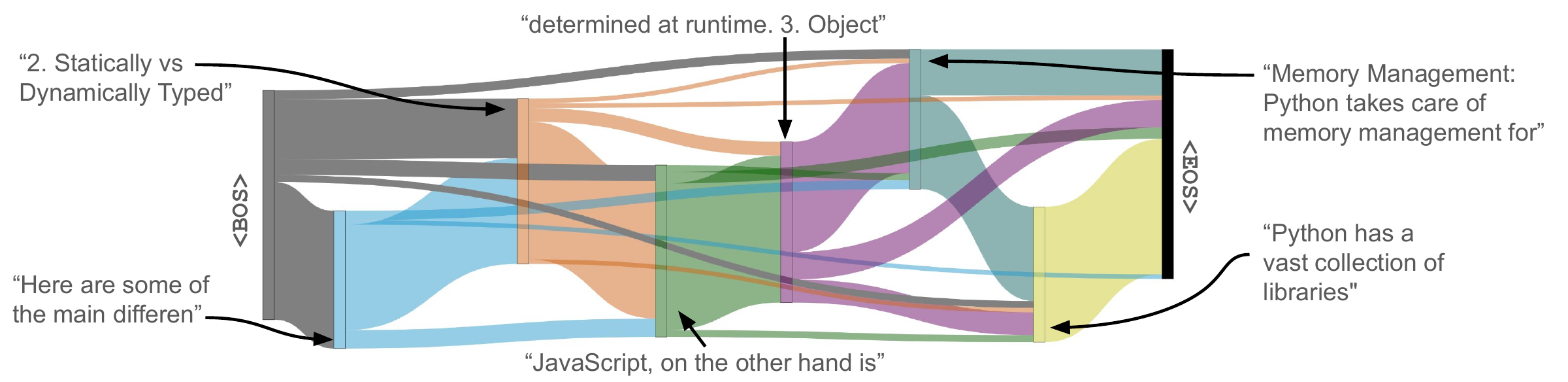}
\end{minipage}
\begin{minipage}{\textwidth} \centering
\includegraphics[width=0.9\linewidth]{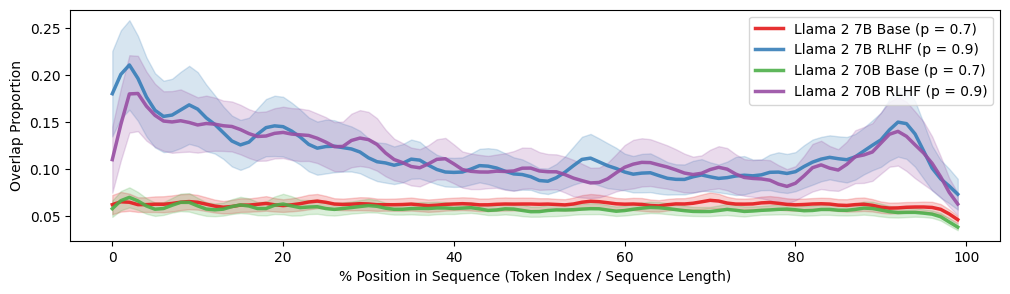}
\end{minipage}

\caption{
\textbf{
RLHF model generations on the same prompt are highly similar to each other, unlike Base LMs.} 
For each of 80 short prompts, we collect and align 100 generations (nucleus sampling, p = 0.9) from Base (pretrained) and RLHF models. \textbf{Above:} (\S\ref{sec:anchor_spans}) A Sankey diagram of 100 RLHF model generations for the prompt ``What are the main differences between Python and JavaScript programming languages?'' Sequences share multiple lengthy anchor spans which appear verbatim in the same order, forming a uniform skeleton for nearly all generations. \textbf{Below:} (\S\ref{sec:backbones}) Over the sequence length, the number of generations aligned with at least 5 others, averaged over all prompts. Base model generations maintain low levels of alignment. RLHF model generations exhibit high alignment throughout, but especially near the beginning and end of generations.
}
\label{fig:fig1}
\end{figure*}

\section{Introduction}
\label{sec:intro}
 {}

Alignment via RLHF \citep{ivison2023camels,touvron2023llama} trains models towards action: completing specific goals and excelling across both short and long-form textual tasks. RLHF works by adapting base LMs that are trained to be \emph{world models}, accurately predicting the distribution of text that might occur after an arbitrary prefix. While RLHF models tend to excel at complex tasks, in this work we find that they partially lose the \emph{world modeling} abilities that allow base LMs to simulate documents from the broader distribution of the internet. We propose that this trade-off is a natural result of RLHF models concentrating probability to specific spans, which allows these models to blueprint long-form generation (Figure~\ref{fig:fig1}) but reduces their ability to model arbitrary text. 

Specifically, RLHF models struggle with next-token prediction which directly measures ability to world model, even when they are finetuned to regain these skills (\S\ref{sec:notllms}). RLHF models seem to concentrate probability on a smaller set of text (\S\ref{sec:collapse}), which follows past work on distributional collapse \citep{shumailov2023curse}. Yet this concentration may have a use: making generation more self-predictable, and helping to blueprint long-form text generations. For example, we find \emph{anchor spans} (Figure~\ref{fig:fig1}) which appear across many samples for the same prompt, and seem to serve as scaffolding for generation (\S\ref{sec:planning}).

Is a trade-off between world modeling and agent modeling fundamental? We argue that self-predictability is likely an inevitable aspect of successful \agents{}, not just a spurious byproduct of RLHF. In order to generate coherent long-form responses (or act towards goals in general) an agent must guarantee that its future actions are largely predictable to its current self. In effect, \emph{agent modeling} may require minimizing long-term uncertainty while \emph{world modeling} requires maintaining the true uncertainty of natural text documents, a fundamental trade-off. We briefly explore this question in \S\ref{sec:tradeoff}. 

Such a trade-off would suggest that methods for adapting models to useful tasks, such as RLHF, will tend to narrow the breadth of a model. In other words, an agent model that takes actions towards a long-term goal may not be fully representative of the broader distribution of all possible agents and goals. General systems covering both sets of abilities might combine agent and world models rather than relying on agent models to both act and predict.



\section{Agent models aren't general language models anymore}
\label{sec:notllms}

Many current language models begin as \emph{world models}, trained to accurately predict the probabilities of possible events in a medium (i.e. text), and are later adapted as \emph{\agents{}}--trained to interact with users towards specific goals. While agent models, particularly those trained with RLHF, are unparalleled as interactive dialogue agents, we show in this section that such adaptation diminishes the original ability to world model, i.e., provide accurate estimated probabilities of text. Our analysis shows that agent models significantly underperform the Base LMs they are trained from on a set of language modeling tasks across diverse domains (\S\ref{sec:ppl}). Even when re-trained towards language modeling (\S\ref{sec: finetune}), they fail to match Base models, suggesting a potential trade-off between the abilities of \emph{agent} and \emph{world} models. 



\begin{figure*}[ht]
\centering
\footnotesize
\begin{minipage}{0.45\linewidth}
    \centering
    \includegraphics[width=\linewidth]{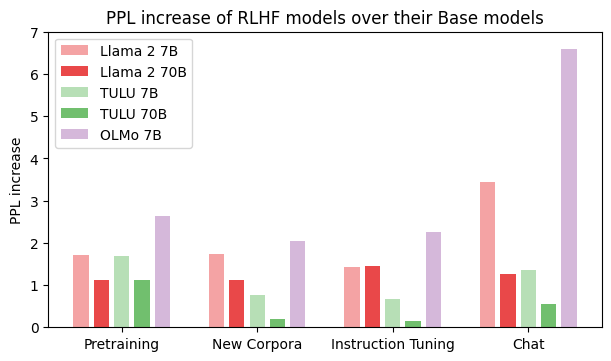}
    \textbf{(A)}
\end{minipage}
\hfill 
\begin{minipage}{0.45\linewidth}
    \centering
    \includegraphics[width=\linewidth]{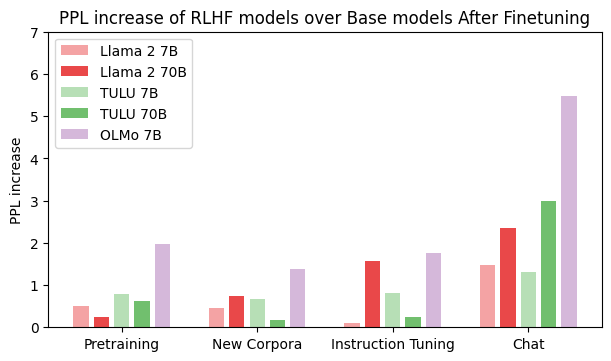}
    \textbf{(B)}
\end{minipage}
\caption{\textbf{(\S\ref{sec:ppl}) RLHF models are significantly worse on language modeling tasks compared to the Base LMs they are adapted from, even on data similar to their preference tuning corpora.}
\textbf{(A):} Perplexity increase in models post-RLHF compared to the base LLM, across several model families and sizes, evaluated on 9 text corpora grouped into 4 categories. RLHF models consistently underperform the Base models they were tuned from. (The lower the perplexity, the better.) \textbf{(B):} We finetune each model from (A) on the target corpus; the general trend remains unchanged. RLHF models are consistently inferior even post-finetuning.  Details in Appendix~\ref{sec:setup-details}.
}
\label{fig:perplexity}
\end{figure*}


\subsection{Perplexity of Base vs. RLHF models} \label{sec:ppl}

The ability to accurately predict the what will happen next given arbitrary starting conditions (i.e., to world model) can be evaluated in the text domain as performance on next-token prediction with the perplexity metric. 
We evaluate the perplexity of the Base models against their \agent{} counterparts. We focus on RLHF as a means of producing agent models, as this is the most popular and effective approach currently being used. Specifically, we analyze two Base models: Llama 2 \citep{touvron2023llama} and OLMo \citep{groeneveld2024olmo}. For their RLHF adaptations, we employ Tulu 2 \citep{ivison2023camels} and Llama 2 Chat \citep{touvron2023llama} for Base Llama 2, and OLMo RLHF for the Base OLMo. We consider common pretraining corpora such as C4 \citep{roberts2019exploring}, Arxiv \citep{clement2019arxiv}, and Wikipedia for evaluation. To test generalization to new data, we also use new corpora released post-model development including new Arxiv papers and BBC news stories~\citep{li2023avoiding}. 
Furthermore, we incorporate instruction finetuning data like Humpback \citep{li2023self} and chat assistant data such as OASST1 \citep{kopf2024openassistant}, Anthropic Harmless and Helpful corpora \citep{bai2022training}. 

Results in Figure~\ref{fig:perplexity}~(A) show that agent models perform consistently worse than the Base models that they were adapted from at language modeling. 
On standard test sets crawled directly from the broader internet (e.g., C4) or specific domains (e.g., Arxiv) this is not a surprising result: RLHF models are trained to produce specific distributions, and thus are no longer good density estimators of the language model pretraining data distribution. However, we show that this holds true \textit{even on instruction tuning and chat assistant datasets} which are exclusively aimed at capturing the action-focused data that current RLHF models are trained towards.

RLHF models are poor zero-shot language models, even on distributions they are trained to imitate, suggesting that they are doing something other than simply capturing these distributions. Indeed, this is not the goal of using RLHF, but it raises a question: what are RLHF models actually doing? In \S\ref{sec:collapse} we argue that RLHF models shift their distribution towards a space that can reliably produce high-reward responses via \textit{implicit blueprints}. This 
limits the generality of predictions from language models adapted to be \agents{}. In other words: adapting models to become \agents{} reduces their capacity as \worldmodels{}. 

\subsection{Readaptation via Finetuning} \label{sec: finetune}

While RLHF certainly warps the distribution of LLMs to be worse predictors, is this merely a surface-level change? Is the next-token-prediction still hidden within the weights of the adapted model? Perhaps the information for arbitrary \prediction{} is simply unused in the output layers of RLHF models.  Fully addressing this concern is beyond the scope of this paper, but we present evidence that it is at least not \textit{trivial} to recover the next-token-prediction capabilities of RLHF models. To test this hypothesis we continue to pretrain both Base models and RLHF models on the training sets of the evaluation corpora used in \S\ref{sec:ppl}. 


The results in Figure~\ref{fig:perplexity}~(B) show that it is difficult to recover the original ability of RLHF models to act as world models. 
Note that this remains true \textit{even for instruction tuning and chat assistant corpora}, which much more closely match the distribution that \agents{} have been adapted to. Further evidence that RLHF models are not even good \textit{rankers} of likely text is shown in Appendix~\ref{app:shannon-game} via the Shannon Game \cite{hovy1998automated}.

\section{Agent models concentrate probability}
\label{sec:collapse}

Past work \citep{shumailov2023curse} has suggested that RLHF models collapse their probability distributions, assigning high probability to a small set of tokens rather than a smoother distribution as observed in Base language models. 
In this section we quantify some properties of this collapse, leading us to propose that RLHF concentrates model probability onto text that is \emph{predictable} by the model, yet still diverse and high quality. This will be an important point in \S\ref{sec:planning} when we discuss how \agents{} appear to have implicit blueprints for generating text by narrowing the scope of possible futures, particularly onto self-predictable text that could make structuring long-form text easier.
\begin{figure}[t]
\centering \footnotesize
\includegraphics[width=0.9\linewidth]{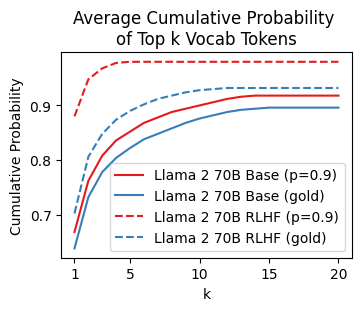}
\caption{
\textbf{(\S\ref{sec:collapse}) RLHF models assign nearly all of the next-token probability mass to a single token, more than Base models.} For Base and RLHF models, we calculate the next-token probability distributions on the gold sequences, as well as on the models' own generations (nucleus sampling, p=0.9). We show the cumulative probability mass of the tokens sorted in descending order of probability. RLHF models assign a larger portion of the probability mass to a small number of tokens, compared to Base models. Details are in Appendix \ref{sec:setup-topk}.
}
\label{fig:cum-probs}
\end{figure}

\begin{figure}[!ht]
\centering \footnotesize
 \includegraphics[width=0.97\linewidth]{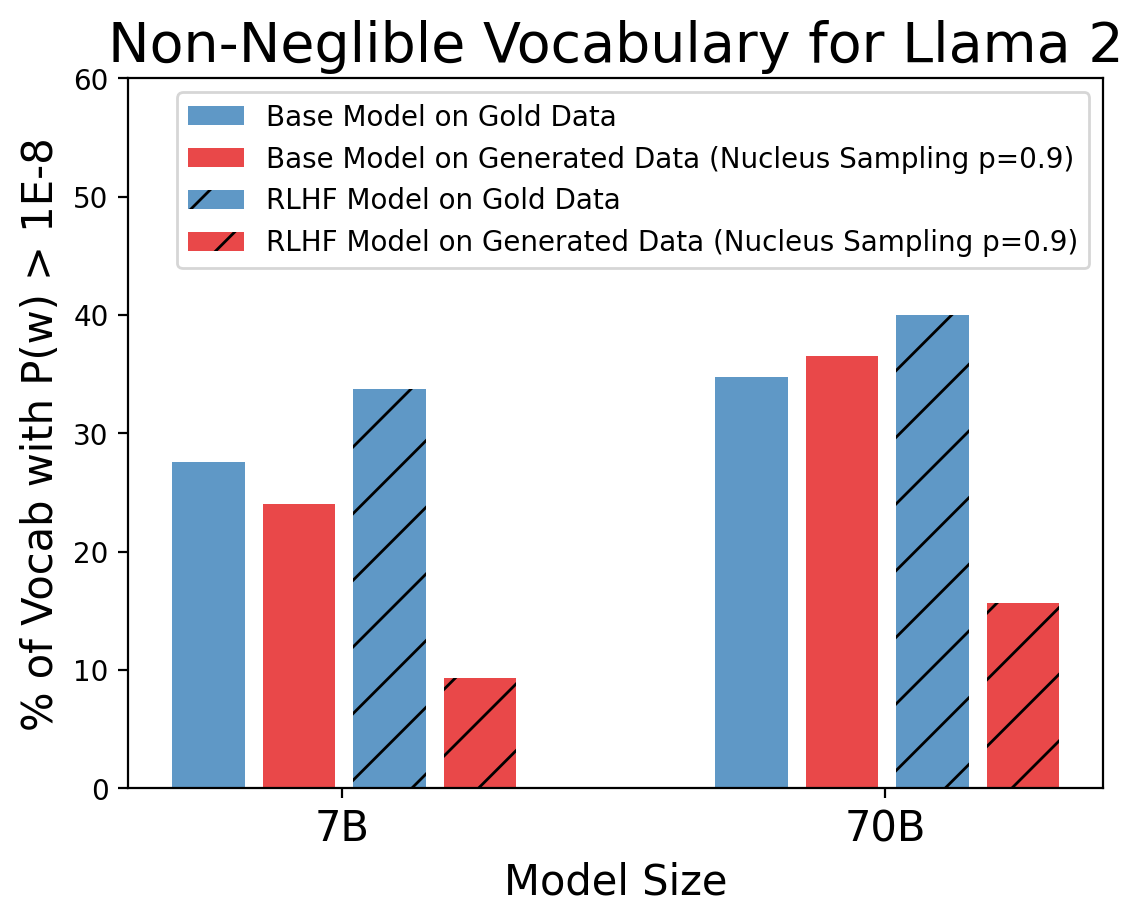}

\caption{
\textbf{(\S\ref{sec:collapse}) RLHF models assign near-zero probability mass to almost all tokens when generating, more than Base models.} For both 7B and 70B models, RLHF models assign non-negligible ($>10^{-8}$) probability to \emph{significantly fewer} vocabulary tokens when predicting next tokens from its own generations (nucleus sampling, p = 0.9), but slightly \emph{more} tokens when predicting next tokens on gold text sequences. This suggests RLHF models only exhibit collapse in their own generative distributions. Details are in Appendix \ref{sec:setup-topk}.
}
\label{fig:non-neg}
\end{figure}


Figure~\ref{fig:cum-probs} shows how concentrated different model distributions are, measured as the probability mass assigned to the most probable $k$ tokens on gold vs. generated data. The most top-heavy distribution is very clearly RLHF on its own sampled generations, e.g., the average probability of the highest probability token in RLHF when conditioning on its own generated text is nearly $0.9$ whereas it is well below $0.7$ for both Base model setups. Interestingly, it is not only for RLHF vs. Base models that we see a noticeable gap, but also RLHF models conditioned on gold vs. generated data. When RLHF models are conditioned on their generated text, they become extremely confident, suggesting that RLHF models remain in a region of confidence once generation has started. 

We can also consider the converse: how \textit{diffuse} model distributions are, measured by how many tokens are assigned non-negligible probability. Figure~\ref{fig:non-neg} displays the average number of vocabulary items with greater than $10^{-8}$ probability per time-step; in other words, the average number of tokens with a non-trivial probability of being sampled at generation time, which we can also think of as a rough measure of generative uncertainty. As the figure shows, RLHF models have vocabulary distributions that tend to be slightly heavier tailed (more uncertain) on human authored data, but much \textit{lighter tailed} (less uncertain) when conditioned on their own generated text. Interestingly, Base models do not differ as drastically between human authored gold data and their own generations, and RLHF on gold data is similar to Base model uncertainty but slightly \textit{more} uncertain than Base models. The fact that RLHF models have such low uncertainty on their own data, but otherwise very high uncertainty, suggests that they are highly self-conditional and adapted to predicting their own distributions,
limiting diversity but allowing for long-term predictability that can aid long-form generation. We explore what this predictability looks like in \S\ref{sec:planning}.

The self-predictability of RLHF models is not merely explained by the low diversity of RLHF models. Adjusting Llama 2 70B RLHF (with nucleus sampling, $p$=0.9) to be roughly as diverse as Llama 2 70B Base (see $n$gram statistics in Table~\ref{tab:ngramdiversity}), we still find Llama 2 70B Base is less confident, even on its own \textit{greedy} generations (see Figure~\ref{fig:cum-probs-with-greedy} in Appendix). In effect, RLHF models seem to be better at self-prediction, even accounting for diversity. 

\begin{figure}[!ht]
\centering \footnotesize

\includegraphics[width=1\linewidth]{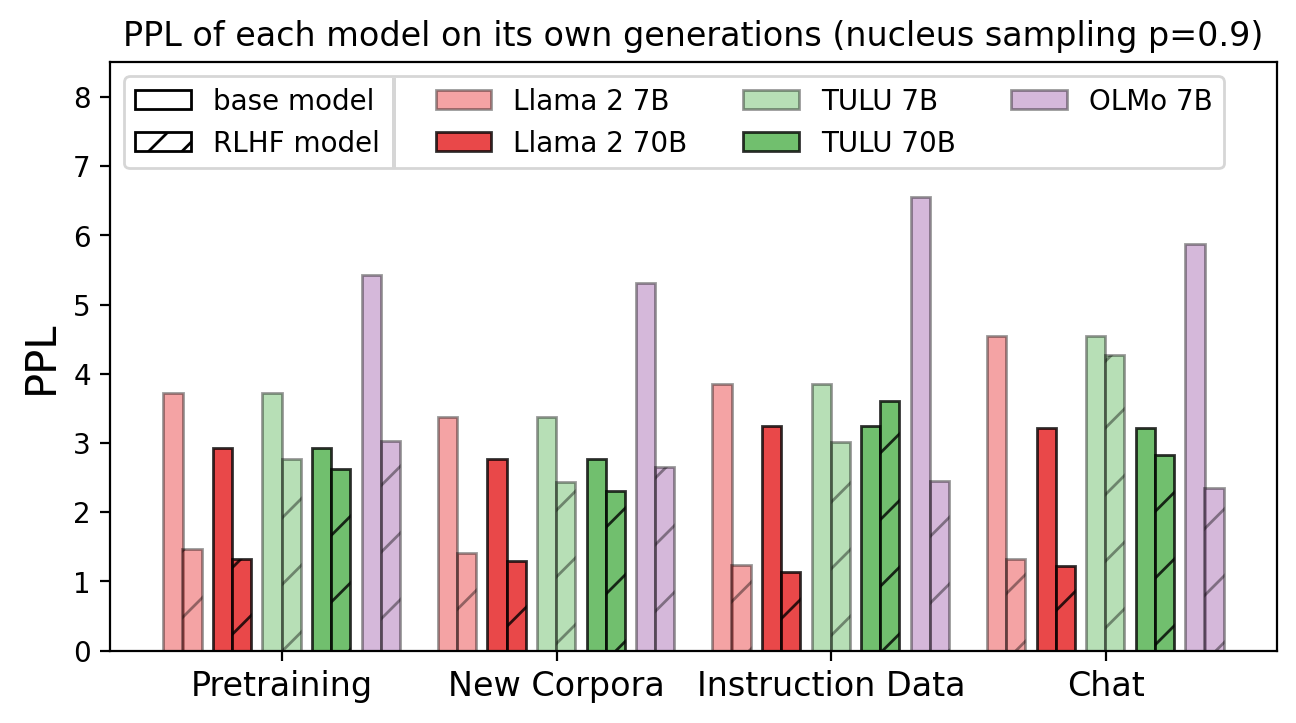}

\caption{
\textbf{(\S\ref{sec:collapse}) RLHF models have lower perplexity when evaluated on their own generations}: We generate (nucleus sampling, p=0.9) completions of prefixes taken from 11 datasets, grouped into 4 categories. 
Details are in Appendix \ref{tbl: self_ppl_data}. 
}
\label{fig:self-perplexity-llama}
\end{figure}

These results suggest that RLHF models may be staying within a region of confidence when they generate text, in turn allowing for better long-form coherency. It seems that \agents{} are significantly better at predicting their own generated text than \predictors{}, which is verified in Figure~\ref{fig:self-perplexity-llama}. This naturally leads to the hypothesis that this kind of high confidence is required to think ahead for long-form generation (\S\ref{sec:planning}), suggesting a deeper tradeoff between \emph{acting} and \emph{predicting}.

\section{Agent models think ahead}
\label{sec:planning}

\begin{figure}[t]
\centering \footnotesize

    \includegraphics[width=\linewidth]{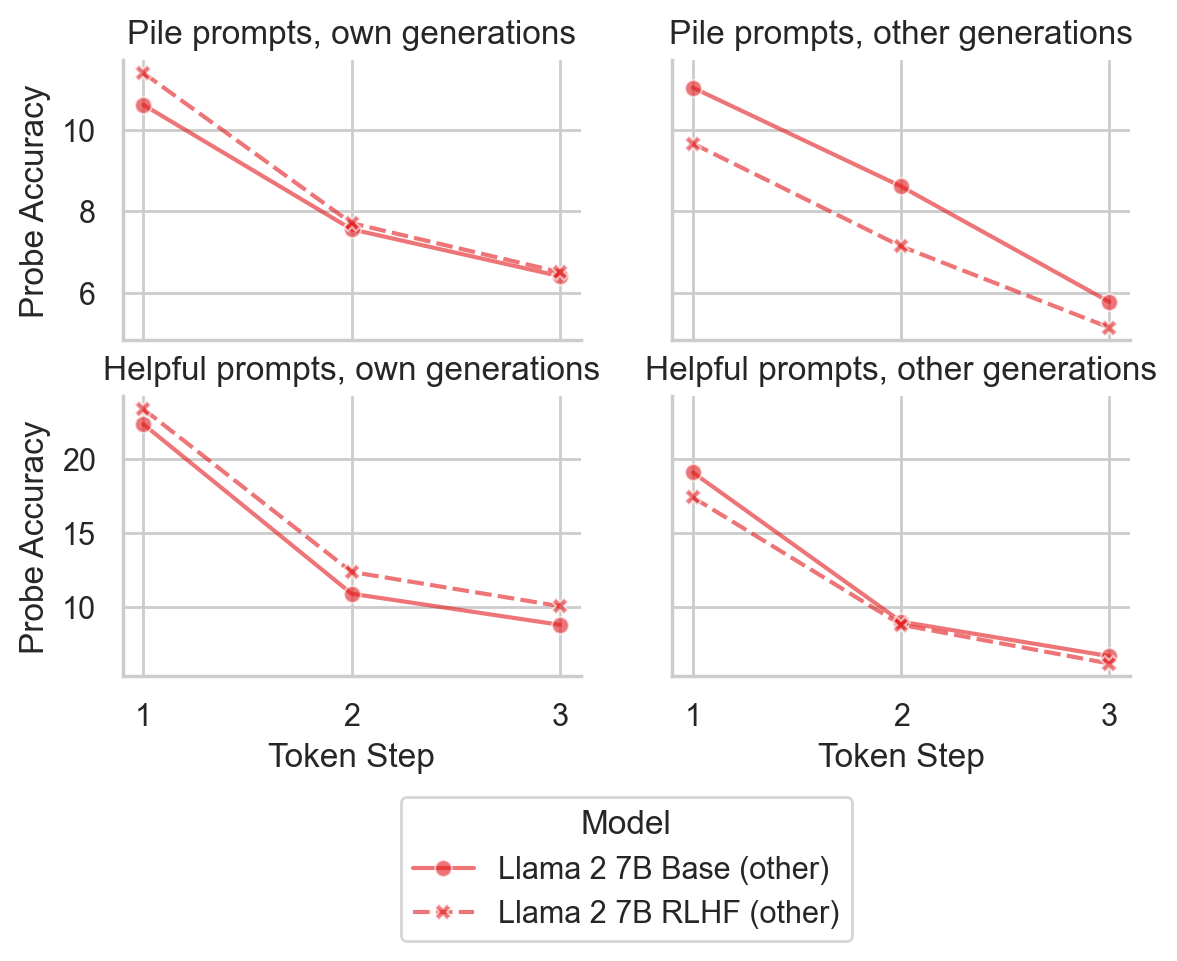}

\caption{
\textbf{
(\S\ref{sec:predict_future}) The intermediate representations of RLHF models contain more linearly extractible information for predicting future tokens of generated text, compared to Base models.}  We train a linear probe to predict each model's future tokens ($n=1,2,3$ tokens into the future) given the hidden representation. Each subplot indicates probe accuracy at predicting the model-generated token $n$ steps in the future.
\textbf{Top} uses prompts from the Pile, and \textbf{Bottom} uses prompts from the Anthropic Helpful dataset. \textbf{Left} predicts the model's own generation, while \textbf{Right} predicts for the other model (e.g. RLHF predicting Base and vice versa).}
\label{fig:probing}
\end{figure}

We have shown that RLHF models are no longer good \predictors{} and that they concentrate their probability distributions into more predictable outcomes. What does this probability mass shift lead to? In this section we show evidence that RLHF models make use of a kind of implicit blueprint for long-form generation rather than predicting only one token in advance, effectively using probability concentrated on certain future spans to enable ``thinking ahead'' in long form generation. 

\subsection{RLHF hidden states are more predictive of future tokens}
\label{sec:predict_future}

As a basic measure of ``thinking ahead'', we first estimate the information models have about future timesteps, beyond the next token. Using the methodology of \citet{pal2023future}, we study how well we can predict tokens for Llama 2 7B Base and RLHF models by leveraging their own hidden states with a linear probe.
Specifically, we perform evaluation on the Pile~\citep{gao2020pile} and Anthropic Helpful~\citep{bai2022training} datasets.
We train a linear model on the hidden representation from the 28th layer (Llama 2 7B comprises 32 layers), using 100,000 tokens from each dataset, to predict tokens $n$ steps into the future ($n=1, 2, 3$), \textit{after} the token being predicted at the current timestep. We use this prediction accuracy as a metric to assess the effectiveness of using a model's hidden state to foresee its own generation of future tokens.


Linear probes on RLHF models can predict future tokens in their own generations with higher accuracy than Base models (Figure~\ref{fig:probing}). RLHF adapted models find it easier to predict the future because they generate a future that is more predictable, as supported by Figure~\ref{fig:self-perplexity-llama}.

Figure~\ref{fig:probing} also shows that RLHF models may contain a subset of information that base LLMs do, as base LLMs are more effective at predicting future tokens of RLHF models than the reverse. This supports the idea of concentration or collapse in RLHF models, as their generations fit Base LLMs, but they no longer explain Base LLM generations as effectively.




\subsection{Sampled generations contain align-able blueprints}
\label{sec:backbones}

\begin{figure*}
\centering \footnotesize

\begin{minipage}{0.48\textwidth}
 \centering
 \includegraphics[width=1\linewidth]{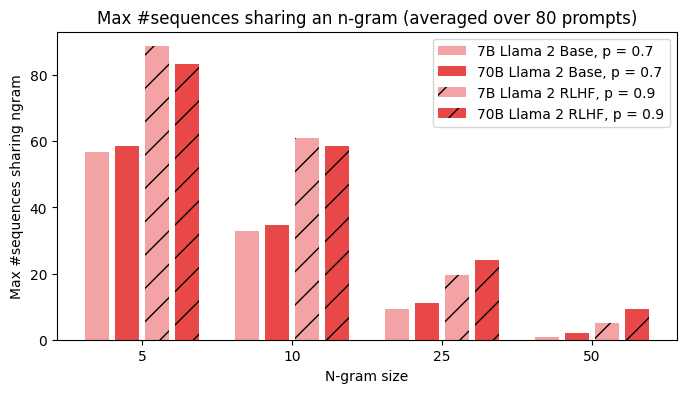}
 \textbf{(A)}
\end{minipage}
\begin {minipage}{0.48\textwidth}
 \centering
 \includegraphics[width=1\linewidth]{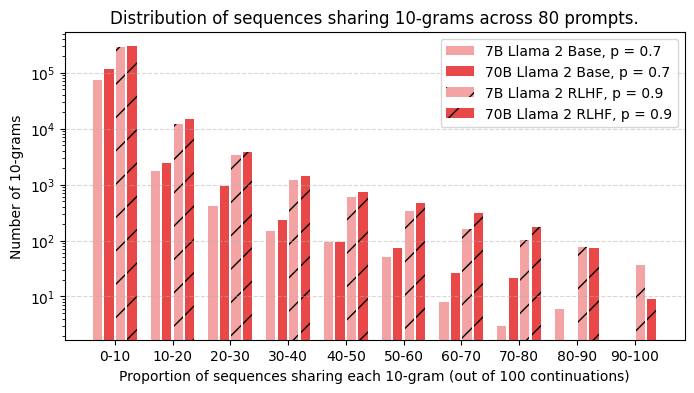}
 \textbf{(B)}
\end{minipage}

\caption{
\textbf{(\S\ref{sec:backbones}) Given a short prompt, RLHF models heavily reuse n-grams across many independently generated continuations (nucleus sampling, p = 0.9), with the most common 10-grams appearing in 60\% of generations on average.} For each of 80 short prompts, we collect 100 generations from Base and RLHF models using nucleus sampling with p = 0.7 and p = 0.9, respectively. \textbf{(A)} For each prompt, the number of generations, out of 100 total, which contain the most common n-gram ($n \in [5, 10, 25, 50]$), averaged across all prompts. \textbf{(B)} A histogram, binning 10-grams by the number of sequences containing that 10-gram. Counts are log-scale. Compared to Base models, RLHF models much more frequently generate the same 10-gram in nearly all continuations for a prompt. Statistics for other models are available in Appendix~\ref{app:ngrams}.
}
\label{fig:ngrams}
\end{figure*}



RLHF models seem to construct implicit blueprints for a given prompt, consistently touching on certain points across different samples. Sampling 100 continuations for the same prompt from an RLHF model leads to sequences with a significant amount of diversity (in terms of unique $n$grams), but a surprisingly high amount of overlap between the sequences. While diversity and overlap may seem to be in conflict, Figure~\ref{fig:ngrams} shows this is not the case. We can use nucleus sampling \citep{holtzman2020curious} to generate text with similar $n$gram diversity across Base LLMs and RLHF-adapted models by setting the value of $p$ appropriately (see Appendix \ref{sec:planning-experiments}). Yet even when diversity statistics are similar, RLHF models reuse more long $n$grams across different generations for the same prompt than models not trained using Reinforcement Learning. In effect, the diversity is not evenly distributed in RLHF generation, with diversity concentrated in between long, predictable spans.

In Figure~\ref{fig:ngrams}, on average a quarter of RLHF outputs share at least one 25-gram (about a sentence), and there are almost twice as many RLHF sequences sharing at least a 10-gram than for the Base model. This high level of \textit{n}gram sharing is not limited to a few isolated cases or to the prefix or suffix of the generations. The bar-chart on the right side of Figure~\ref{fig:ngrams}~(B) shows that on average 10-grams are shared by more sequences in RLHF model outputs by a large margin (note log-scale).

However, \textit{n}gram statistics are not robust to small variations, nor do they consider ordering effects—where \textit{n}grams in RLHF models tend to occur in certain orders. 
To handle these problems, we propose to first align the continuations for a given prompt using a sequence alignment method  originally developed for bioinformatics, MAFFT \citep{katoh2013mafft}.

After aligning 100 continuations for each prompt (Figure~\ref{fig:fig1}~bottom), Base models have a much lower overlap compared to RLHF models despite having controlled for \textit{n}gram diversity using the nucleus sampling parameter $p$, as described in Appendix~\ref{sec:planning-experiments}. RLHF models have significantly more overlap for all positions, and at the beginning and end of the sequences the overlap is even more drastic. This shows that RLHF models tend to converge back towards the end of the generations. Additionally, we note that the larger standard deviations (shaded in the figure) result from the fact that the convergence points occur at different positions in the sequences for different prompts. These results hold true across different models, see Appendix~\ref{app:overlap}.

\begin{figure*}[t!h]
\centering \footnotesize

\begin {minipage}{\textwidth} \centering
\includegraphics[width=1\linewidth]{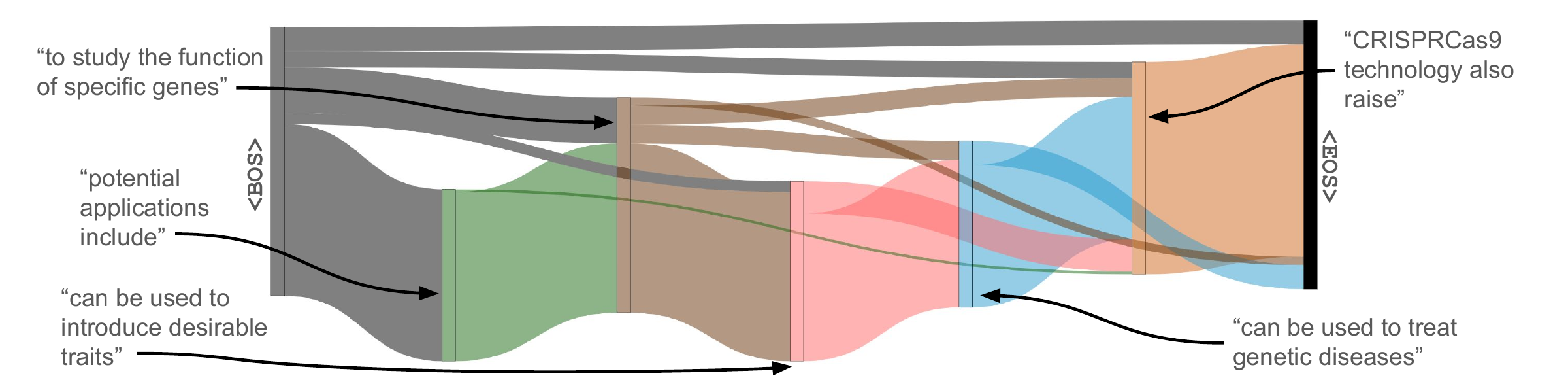}
 \textbf{Llama 2 70B RLHF, Nucleus Sampling, $p$=0.9}
\end{minipage}
\begin{minipage}{\textwidth} \centering
\includegraphics[width=1\linewidth]{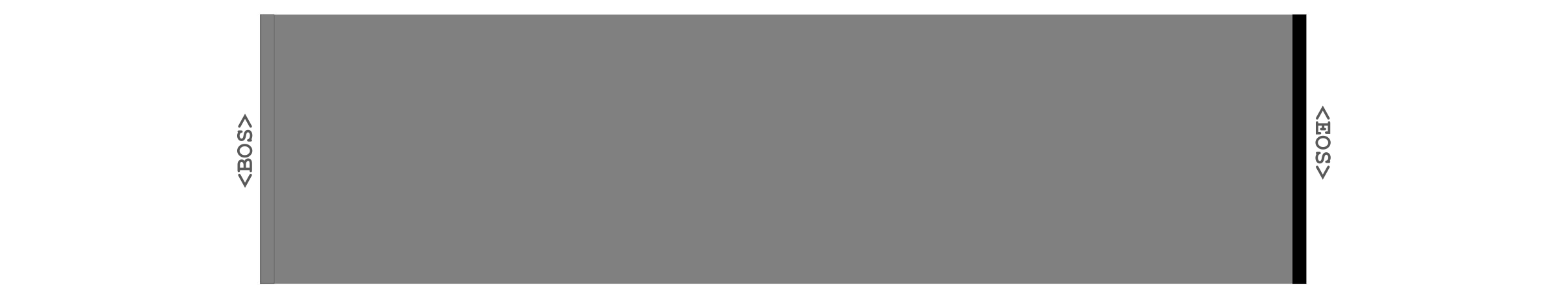}
 \textbf{Llama 2 70B Base, Nucleus Sampling, $p$=0.7}
\end{minipage}

\caption{
  \textbf{(\S\ref{sec:anchor_spans}) The RLHF model (top) shows multiple anchor spans appearing verbatim in the same order, whereas the Base model (bottom) lacks any common anchored points, highlighting the divergence of its generated responses.} Each Sankey diagram visualizes 100 responses (nucleus sampling, $p=0.9$ for RLHF and $p=0.7$ for Base) to the prompt ``Explain the process of gene editing using CRISPR-Cas9 technology, and discuss its potential applications and ethical implications.''
}
\label{fig:sankey}
\end{figure*}

\subsection{Anchor spans as an implicit blueprint}
\label{sec:anchor_spans}

Visualizing these alignments suggests that RLHF models seem to \emph{blueprint} generations, consistently using the same key points across many generations for a given prompt (Figure~\ref{fig:fig1}~top). Base LLMs show no such structure (Figure~\ref{fig:sankey}~bottom).

We visualize these blueprints with Sankey diagrams (Figure~\ref{fig:fig1}~top), which characterize the flow through critical nodes over a sequential process. We focus on  ``anchor spans''—substrings that occur across many sampled outputs for the same prompt—such that the flow between two nodes $A$ and $B$ represents the set of generations which contain $A$, followed by $B$, with (possibly different) text in between. Specifically, we first identify all text spans of a fixed minimum length (30 characters) that occur at the same index in the alignments found in \S\ref{sec:backbones}, for at least some threshold number of generations (20\%). If the same span occurs at two different positions in the alignment indexing, they are counted as different spans. We sort the spans in descending order of frequency across unique outputs, and then by span length. We then greedily pick up to a maximum (6 in these visualizations) number of spans, updating the occurrence counts for unpicked spans after each addition, so that generations cannot ``double count'' for spans which overlap. See Appendix \ref{sec:planning-experiments} for further details.

RLHF models seem to blueprint their long-form generations: Figure~\ref{fig:sankey}~(top) shows that anchor spans remain constant across many outputs, analogous to a bullet point outline embedded within a generation.  For the same prompt the Base model (Llama 2 70B) has no such structure whatsoever, even when using a more restricted sampling strategy (Figure~\ref{fig:sankey}~bottom). In other words, 
RLHF models produce an implicit blueprint, converging to certain spans from which they can reliably predict their own future, while Base models diverge in unpredictable ways (more examples Appendix~\ref{sec:appendix-sankey}), even when adjusting for diversity.

\section{Is this a fundamental trade-off?}
\label{sec:tradeoff}

Are these differences in \worldmodels{} and \agents{} fundamental, or just a result of current language model training and RLHF practices? Even if new agent adaptation methods alleviate some of these problems, we propose that, under fixed capacity, there must necessarily be some trade-off between \worldmodels{} and \agents{} because of the underlying differences in their optimal behavior, which we are more closely approaching with recent state-of-the-art methods. 

\subsection{Why can't world models just be agent models when we ask them to be?}

Prompts almost never fully specify desired behavior. Base LLMs have been trained to sample from the space of \textit{possible documents with a given prefix}. In the overwhelming majority of cases, prefixes do not fully specify many of the choices a document can represent.  Furthermore, Base LLMs have been shown to be very sensitive to short-term context \citep{paperno-EtAl:2016:P16-1, wang2023mint}. This is a feature, not a bug, of \worldmodels{}, as short-term context is generally more predictive than long-term context in human-authored documents. Yet it means that Base LLMs are highly sensitive to sampling even one incoherent token. \citet{dziri2024faith} suggest that the probability of generating an error is lower bounded by the $1 - P(\textrm{error})^\ell$ where $\ell$ is the length of a document. The actual error rate is likely much higher, as the model conditions on previously made errors, creating a snowball effect, as with hallucination \citep{zhang2023language}.

\subsection{Planning around randomness}
\label{ssec:planning-around-randomness}
The ability for agents to plan well is directly related to the amount of randomness in their environment. Reinforcement Learning (RL) is the mathematical language of agent modeling, so we take a moment to conceptualize why RLHF models would collapse and rely on anchor spans in terms of RL. 
Recall the Bellman equation, describing the value of a current state of a sequence of actions:
\begin{equation}
V(s) = \max_{a \in A} \left\{ R(s, a) + \gamma \sum_{s' \in S} P(s' | s, a) V(s') \right\}
\end{equation}
where $V$ is the value of state $s$, $A$ is the set of possible actions, $R$ is the reward function, $0 < \gamma < 1$ is the discount factor, $S$ is set of all possible states, and $P$ is the transition function. Note that if $P$ is uniform over $S$ at every state regardless of action, there is no need for planning at all: in a purely stochastic environment, planning is impossible. This holds true even if for a given state $s$ there are a limited subset of states $S_{s\rightarrow} \subset S$ that have non-zero probability under $P$, because $P(s'|s,a_1) = P(s'|s,a_2)$ in all of these cases, i.e., the action an agent chooses to take has no effect on the outcome, so planning is entirely unnecessary.

On the other extreme, if an agent can freely choose their next state, e.g., $\forall s \in S, \forall s' \in S_{s\rightarrow}, \exists a_{s'}: P(s'|s,a_{s'}) = 1$, where $S_{s\rightarrow}$ is the set of reachable states from $s$, it will always do as well or better than an environment with the same connectivity but increased stochasticity:
\begin{equation}
\label{eqn:stoch-is-worse}
\max_{a' \in A} \sum_{s' \in S}  P(s' | s, a) R(s', a') \leq \max_{a' \in A} R(s', a') 
\end{equation}
Holding all else equal, a stochastic transition function means that even an optimal agent will sometimes enter a suboptimal solution due to incomplete control over future states.

Base LLMs adapted with RLHF face a similar conundrum to a highly stochastic transition function: they are forced to learn with a highly stochastic initial policy $\pi_{\textrm{Base LLM}}$
\vspace{3pt}
in an environment with a huge space of states and non-smooth distribution of rewards. Models therefore may tend towards previously high-reward states (e.g. the \nervecenterspans{}), as this is less risky than exploration. In the exponentially large space of possible strings it is unsurprising that models tend to converge towards \nervecenterspans{}.

Adapting LLMs for long-form generation incentivizes the use of anchor spans, in order to avoid entering states for which $\pi_{\textrm{Base LLM}}$ has high entropy and is therefore hard to plan around. This becomes exponentially harder to avoid as the length of generated strings grows \citep{dziri2024faith}. We suggest that models converge to anchor spans to lower-bound success, and will likely do so when adapted from a vanilla next-token-predictor, even with very different adaptation methods.

\section{Related Work}


\paragraph{Catastrophic Forgetting} Prior work examined models forgetting previously learned distributions. In most continual learning settings, where the model continues to train on new data, this is termed \emph{catastrophic forgetting} \citep{kirkpatrick2017overcoming}. This phenomenon has been observed in Language Models, and several mechanisms have been proposed to mitigate its effects \citep{chen2020recall,xu2020forget,vu2022overcoming}. More recently, catastrophic forgetting has also been found to impact modern generative LLMs \citep{luo2023empirical}.

\paragraph{Distribution collapse of LLMs} Distribution collapse is known to occur in models which have been trained on data distributions that include model generations \citep{shumailov2023curse}. Such self-consuming models exhibit a degradation in generation quality as well as diversity \citep{briesch2023large,alemohammad2023self}. Unlike these works, which do not consider preference tuning objectives at all, we focus on collapse in RLHF models. 
Other works do consider RLHF models, but do not study the nature of this collapse, instead focusing exclusively on methods for alleviating the style of \emph{mode collapse} which arises from the degenerative overfitting to an imperfect reward model \citep{perez2022red,go2023aligning}. These methods are taken from the RL literature \citep{pmlr-v70-jaques17a}, and build on studies which find mode collapse common in many other models \citep{che2017mode,pmlr-v80-mescheder18a}.

Concurrent with our work, \citet{Lake2024-kf} show that RLHF appears to be collapsing onto a subset of the Base LLM distribution, as we hypothesize in \S\ref{sec:predict_future}, suggesting that RLHF has lower diversity due to an aggregating effect that can be partially (but not fully) mimicked with prompting. \citet{xiao2024algorithmic} also studies RLHF models, specifically on collapse at a \emph{preference} level over all generations. Our work, on the other hand, visualizes and describes, quantitatively and qualitatively, the sequence-level multi-token repetition in generations after RLHF, as characterized by the presence of \emph{anchor spans}. 

\section{Conclusion}

We present evidence that: (1) RLHF-adapted LLMs are no longer performant next-token predictors, and thus no longer serve as world models of the textual space. (2) Such models concentrate their probability into a more predictable subdistribution of the Base model. (3) RLHF LLMs consistently produce \nervecenterspans{} which can serve as a \textit{blueprint} for long-form generations.

We propose that these differences may indicate a deeper trade-off between \worldmodels{} and \agents{}.
An agent model that takes action via sampling might reshape its distribution in such a way that it no longer represents the full set of possibilities that \worldmodels{} capture, ensuring that it doesn't ``go off the rails'' as Base LLMs are prone to.
Future work could explore strategies that can mitigate the observed trade-off between the predictive capabilities of world models and the action-oriented nature of agent models. For instance, a system that decides when to call a \worldmodel{} vs. an \agent{} may be able to more reliably switch between these for different goals, using the \worldmodel{} as a probabilistic simulator and predictor that helps the \agent{} decide how to act.


\vfill
\pagebreak

\section*{Limitations}

One key limitation of our work that will need to be addressed in future literature is comprehensiveness across a broader range of models, as well as varied methods for agent-alignment. We focus on RLHF here as this is the most popular and successful method currently being used. We also aimed to test popular and performant models, as these represent the limits of both RLHF and base LM capabilities. In future work, it will be important to study more various models, including more model scales. It will also be useful to include multiple random seeds for training/tuning in all experiments, but the cost of such experiments would be infeasible here.

It will also be useful in future work to study the blueprints produced by RLHF models in more detail. This work aims to both demonstrate the trade-off between agent and world modeling, then explore different aspects of this blueprinting. However, dedicated work will be required to fully characterize this process.

\section*{Ethical Considerations}

Aligned models, such as those tuned with RLHF, have seen a recent explosion in capabilities, popularity, and deployment compared to traditional LLMs trained primarily on broad text prediction. As part of this shift, aligned models are much more frequently framed as ``agents'' which can take explicit actions, and take active part in human-facing systems. This framing poses significant risks without a better conceptual understanding of these techniques, particularly with respect to whether these models are indeed robust agents and what underlying learned mechanisms might allow for this. 

Our work seeks to contribute to this understanding. For example, we find some evidence that aligned models may indeed be planning, which supports a notion of ``agentiveness''. Yet we also find that these models lose robust and accessible notions of calibrated text prediction, which could indicate a tendency towards biased heuristics or at least away from the more robust, broad-text understanding of traditional LMs. Overall, our work indicates ongoing concerns with treating aligned models as well-informed agents, and demands further study into these aspects and risks.

\bibliography{custom}

\begin{thebibliography}{35}
\providecommand{\natexlab}[1]{#1}

\bibitem[{Alemohammad et~al.(2023)Alemohammad, Casco-Rodriguez, Luzi, Humayun, Babaei, LeJeune, Siahkoohi, and Baraniuk}]{alemohammad2023self}
Sina Alemohammad, Josue Casco-Rodriguez, Lorenzo Luzi, Ahmed~Imtiaz Humayun, Hossein Babaei, Daniel LeJeune, Ali Siahkoohi, and Richard~G Baraniuk. 2023.
\newblock Self-consuming generative models go mad.
\newblock \emph{arXiv preprint arXiv:2307.01850}.

\bibitem[{Bai et~al.(2022)Bai, Jones, Ndousse, Askell, Chen, DasSarma, Drain, Fort, Ganguli, Henighan et~al.}]{bai2022training}
Yuntao Bai, Andy Jones, Kamal Ndousse, Amanda Askell, Anna Chen, Nova DasSarma, Dawn Drain, Stanislav Fort, Deep Ganguli, Tom Henighan, et~al. 2022.
\newblock Training a helpful and harmless assistant with reinforcement learning from human feedback.
\newblock \emph{arXiv preprint arXiv:2204.05862}.

\bibitem[{Briesch et~al.(2023)Briesch, Sobania, and Rothlauf}]{briesch2023large}
Martin Briesch, Dominik Sobania, and Franz Rothlauf. 2023.
\newblock Large language models suffer from their own output: An analysis of the self-consuming training loop.
\newblock \emph{arXiv preprint arXiv:2311.16822}.

\bibitem[{Che et~al.(2017)Che, Li, Jacob, Bengio, and Li}]{che2017mode}
Tong Che, Yanran Li, Athul Jacob, Yoshua Bengio, and Wenjie Li. 2017.
\newblock \href {https://openreview.net/forum?id=HJKkY35le} {Mode regularized generative adversarial networks}.
\newblock In \emph{International Conference on Learning Representations}.

\bibitem[{Chen et~al.(2020)Chen, Hou, Cui, Che, Liu, and Yu}]{chen2020recall}
Sanyuan Chen, Yutai Hou, Yiming Cui, Wanxiang Che, Ting Liu, and Xiangzhan Yu. 2020.
\newblock Recall and learn: Fine-tuning deep pretrained language models with less forgetting.
\newblock \emph{arXiv preprint arXiv:2004.12651}.

\bibitem[{Chiang et~al.(2023)Chiang, Li, Lin, Sheng, Wu, Zhang, Zheng, Zhuang, Zhuang, Gonzalez, Stoica, and Xing}]{vicuna2023}
Wei-Lin Chiang, Zhuohan Li, Zi~Lin, Ying Sheng, Zhanghao Wu, Hao Zhang, Lianmin Zheng, Siyuan Zhuang, Yonghao Zhuang, Joseph~E. Gonzalez, Ion Stoica, and Eric~P. Xing. 2023.
\newblock \href {https://lmsys.org/blog/2023-03-30-vicuna/} {Vicuna: An open-source chatbot impressing gpt-4 with 90\%* chatgpt quality}.

\bibitem[{Clement et~al.(2019)Clement, Bierbaum, O'Keeffe, and Alemi}]{clement2019arxiv}
Colin~B. Clement, Matthew Bierbaum, Kevin~P. O'Keeffe, and Alexander~A. Alemi. 2019.
\newblock \href {https://arxiv.org/abs/1905.00075} {On the use of arxiv as a dataset}.
\newblock \emph{Preprint}, arXiv:1905.00075.

\bibitem[{Dziri et~al.(2024)Dziri, Lu, Sclar, Li, Jiang, Lin, Welleck, West, Bhagavatula, Le~Bras et~al.}]{dziri2024faith}
Nouha Dziri, Ximing Lu, Melanie Sclar, Xiang~Lorraine Li, Liwei Jiang, Bill~Yuchen Lin, Sean Welleck, Peter West, Chandra Bhagavatula, Ronan Le~Bras, et~al. 2024.
\newblock Faith and fate: Limits of transformers on compositionality.
\newblock \emph{Advances in Neural Information Processing Systems}, 36.

\bibitem[{Gao et~al.(2020)Gao, Biderman, Black, Golding, Hoppe, Foster, Phang, He, Thite, Nabeshima et~al.}]{gao2020pile}
Leo Gao, Stella Biderman, Sid Black, Laurence Golding, Travis Hoppe, Charles Foster, Jason Phang, Horace He, Anish Thite, Noa Nabeshima, et~al. 2020.
\newblock The {P}ile: An 800{GB} dataset of diverse text for language modeling.
\newblock \emph{arXiv preprint arXiv:2101.00027}.

\bibitem[{Go et~al.(2023)Go, Korbak, Kruszewski, Rozen, Ryu, and Dymetman}]{go2023aligning}
Dongyoung Go, Tomasz Korbak, Germán Kruszewski, Jos Rozen, Nahyeon Ryu, and Marc Dymetman. 2023.
\newblock \href {https://arxiv.org/abs/2302.08215} {Aligning language models with preferences through f-divergence minimization}.
\newblock \emph{Preprint}, arXiv:2302.08215.

\bibitem[{Groeneveld et~al.(2024)Groeneveld, Beltagy, Walsh, Bhagia, Kinney, Tafjord, Jha, Ivison, Magnusson, Wang et~al.}]{groeneveld2024olmo}
Dirk Groeneveld, Iz~Beltagy, Pete Walsh, Akshita Bhagia, Rodney Kinney, Oyvind Tafjord, Ananya~Harsh Jha, Hamish Ivison, Ian Magnusson, Yizhong Wang, et~al. 2024.
\newblock Olmo: Accelerating the science of language models.
\newblock \emph{arXiv preprint arXiv:2402.00838}.

\bibitem[{Holtzman et~al.(2020)Holtzman, Buys, Du, Forbes, and Choi}]{holtzman2020curious}
Ari Holtzman, Jan Buys, Li~Du, Maxwell Forbes, and Yejin Choi. 2020.
\newblock The curious case of neural text degeneration.
\newblock In \emph{International Conference on Learning Representations}.

\bibitem[{Hovy and Lin(1998)}]{hovy1998automated}
Eduard Hovy and Chin-Yew Lin. 1998.
\newblock Automated text summarization and the summarist system.
\newblock In \emph{TIPSTER TEXT PROGRAM PHASE III: Proceedings of a Workshop held at Baltimore, Maryland, October 13-15, 1998}, pages 197--214.

\bibitem[{Ivison et~al.(2023)Ivison, Wang, Pyatkin, Lambert, Peters, Dasigi, Jang, Wadden, Smith, Beltagy et~al.}]{ivison2023camels}
Hamish Ivison, Yizhong Wang, Valentina Pyatkin, Nathan Lambert, Matthew Peters, Pradeep Dasigi, Joel Jang, David Wadden, Noah~A Smith, Iz~Beltagy, et~al. 2023.
\newblock Camels in a changing climate: Enhancing lm adaptation with tulu 2.
\newblock \emph{arXiv preprint arXiv:2311.10702}.

\bibitem[{Jaques et~al.(2017)Jaques, Gu, Bahdanau, Hern{\'a}ndez-Lobato, Turner, and Eck}]{pmlr-v70-jaques17a}
Natasha Jaques, Shixiang Gu, Dzmitry Bahdanau, Jos{\'e}~Miguel Hern{\'a}ndez-Lobato, Richard~E. Turner, and Douglas Eck. 2017.
\newblock \href {https://proceedings.mlr.press/v70/jaques17a.html} {Sequence tutor: Conservative fine-tuning of sequence generation models with {KL}-control}.
\newblock In \emph{Proceedings of the 34th International Conference on Machine Learning}, volume~70 of \emph{Proceedings of Machine Learning Research}, pages 1645--1654. PMLR.

\bibitem[{Katoh and Standley(2013)}]{katoh2013mafft}
Kazutaka Katoh and Daron~M Standley. 2013.
\newblock Mafft multiple sequence alignment software version 7: improvements in performance and usability.
\newblock \emph{Molecular biology and evolution}, 30(4):772--780.

\bibitem[{Kirkpatrick et~al.(2017)Kirkpatrick, Pascanu, Rabinowitz, Veness, Desjardins, Rusu, Milan, Quan, Ramalho, Grabska-Barwinska et~al.}]{kirkpatrick2017overcoming}
James Kirkpatrick, Razvan Pascanu, Neil Rabinowitz, Joel Veness, Guillaume Desjardins, Andrei~A Rusu, Kieran Milan, John Quan, Tiago Ramalho, Agnieszka Grabska-Barwinska, et~al. 2017.
\newblock Overcoming catastrophic forgetting in neural networks.
\newblock \emph{Proceedings of the national academy of sciences}, 114(13):3521--3526.

\bibitem[{K{\"o}pf et~al.(2024)K{\"o}pf, Kilcher, von R{\"u}tte, Anagnostidis, Tam, Stevens, Barhoum, Nguyen, Stanley, Nagyfi et~al.}]{kopf2024openassistant}
Andreas K{\"o}pf, Yannic Kilcher, Dimitri von R{\"u}tte, Sotiris Anagnostidis, Zhi~Rui Tam, Keith Stevens, Abdullah Barhoum, Duc Nguyen, Oliver Stanley, Rich{\'a}rd Nagyfi, et~al. 2024.
\newblock Openassistant conversations-democratizing large language model alignment.
\newblock \emph{Advances in Neural Information Processing Systems}, 36.

\bibitem[{Lake et~al.(2024)Lake, Choi, and Durrett}]{Lake2024-kf}
Thom Lake, Eunsol Choi, and Greg Durrett. 2024.
\newblock From distributional to overton pluralism: Investigating large language model alignment.
\newblock \emph{arXiv [cs.CL]}.

\bibitem[{Li et~al.(2023{\natexlab{a}})Li, Yu, Zhou, Schick, Levy, Zettlemoyer, Weston, and Lewis}]{li2023self}
Xian Li, Ping Yu, Chunting Zhou, Timo Schick, Omer Levy, Luke Zettlemoyer, Jason~E Weston, and Mike Lewis. 2023{\natexlab{a}}.
\newblock Self-alignment with instruction backtranslation.
\newblock In \emph{The Twelfth International Conference on Learning Representations}.

\bibitem[{Li et~al.(2023{\natexlab{b}})Li, Guerin, and Lin}]{li2023avoiding}
Yucheng Li, Frank Guerin, and Chenghua Lin. 2023{\natexlab{b}}.
\newblock \href {https://arxiv.org/abs/2312.12343} {Avoiding data contamination in language model evaluation: Dynamic test construction with latest materials}.
\newblock \emph{Preprint}, arXiv:2312.12343.

\bibitem[{Luo et~al.(2023)Luo, Yang, Meng, Li, Zhou, and Zhang}]{luo2023empirical}
Yun Luo, Zhen Yang, Fandong Meng, Yafu Li, Jie Zhou, and Yue Zhang. 2023.
\newblock An empirical study of catastrophic forgetting in large language models during continual fine-tuning.
\newblock \emph{arXiv preprint arXiv:2308.08747}.

\bibitem[{Mescheder et~al.(2018)Mescheder, Geiger, and Nowozin}]{pmlr-v80-mescheder18a}
Lars Mescheder, Andreas Geiger, and Sebastian Nowozin. 2018.
\newblock \href {https://proceedings.mlr.press/v80/mescheder18a.html} {Which training methods for {GAN}s do actually converge?}
\newblock In \emph{Proceedings of the 35th International Conference on Machine Learning}, volume~80 of \emph{Proceedings of Machine Learning Research}, pages 3481--3490. PMLR.

\bibitem[{Pal et~al.(2023)Pal, Sun, Yuan, Wallace, and Bau}]{pal2023future}
Koyena Pal, Jiuding Sun, Andrew Yuan, Byron~C Wallace, and David Bau. 2023.
\newblock Future lens: Anticipating subsequent tokens from a single hidden state.
\newblock In \emph{Proceedings of the 27th Conference on Computational Natural Language Learning (CoNLL)}, pages 548--560.

\bibitem[{Paperno et~al.(2016)Paperno, Kruszewski, Lazaridou, Pham, Bernardi, Pezzelle, Baroni, Boleda, and Fernandez}]{paperno-EtAl:2016:P16-1}
Denis Paperno, Germ\'{a}n Kruszewski, Angeliki Lazaridou, Ngoc~Quan Pham, Raffaella Bernardi, Sandro Pezzelle, Marco Baroni, Gemma Boleda, and Raquel Fernandez. 2016.
\newblock \href {http://www.aclweb.org/anthology/P16-1144} {The {LAMBADA} dataset: Word prediction requiring a broad discourse context}.
\newblock In \emph{Proceedings of the 54th Annual Meeting of the Association for Computational Linguistics (Volume 1: Long Papers)}, pages 1525--1534, Berlin, Germany. Association for Computational Linguistics.

\bibitem[{Perez et~al.(2022)Perez, Huang, Song, Cai, Ring, Aslanides, Glaese, McAleese, and Irving}]{perez2022red}
Ethan Perez, Saffron Huang, Francis Song, Trevor Cai, Roman Ring, John Aslanides, Amelia Glaese, Nat McAleese, and Geoffrey Irving. 2022.
\newblock \href {https://arxiv.org/abs/2202.03286} {Red teaming language models with language models}.
\newblock \emph{Preprint}, arXiv:2202.03286.

\bibitem[{Roberts et~al.(2019)Roberts, Raffel, Lee, Matena, Shazeer, Liu, Narang, Li, and Zhou}]{roberts2019exploring}
Adam Roberts, Colin Raffel, Katherine Lee, Michael Matena, Noam Shazeer, Peter~J. Liu, Sharan Narang, Wei Li, and Yanqi Zhou. 2019.
\newblock Exploring the limits of transfer learning with a unified text-to-text transformer.
\newblock Technical report, Google.

\bibitem[{Shumailov et~al.(2023)Shumailov, Shumaylov, Zhao, Gal, Papernot, and Anderson}]{shumailov2023curse}
Ilia Shumailov, Zakhar Shumaylov, Yiren Zhao, Yarin Gal, Nicolas Papernot, and Ross Anderson. 2023.
\newblock The curse of recursion: Training on generated data makes models forget.
\newblock \emph{arXiv preprint arXiv:2305.17493}.

\bibitem[{Touvron et~al.(2023)Touvron, Martin, Stone, Albert, Almahairi, Babaei, Bashlykov, Batra, Bhargava, Bhosale et~al.}]{touvron2023llama}
Hugo Touvron, Louis Martin, Kevin Stone, Peter Albert, Amjad Almahairi, Yasmine Babaei, Nikolay Bashlykov, Soumya Batra, Prajjwal Bhargava, Shruti Bhosale, et~al. 2023.
\newblock Llama 2: Open foundation and fine-tuned chat models.
\newblock \emph{arXiv preprint arXiv:2307.09288}.

\bibitem[{Vu et~al.(2022)Vu, Barua, Lester, Cer, Iyyer, and Constant}]{vu2022overcoming}
Tu~Vu, Aditya Barua, Brian Lester, Daniel Cer, Mohit Iyyer, and Noah Constant. 2022.
\newblock Overcoming catastrophic forgetting in zero-shot cross-lingual generation.
\newblock \emph{arXiv preprint arXiv:2205.12647}.

\bibitem[{Wang et~al.(2023)Wang, Wang, Liu, Chen, Yuan, Peng, and Ji}]{wang2023mint}
Xingyao Wang, Zihan Wang, Jiateng Liu, Yangyi Chen, Lifan Yuan, Hao Peng, and Heng Ji. 2023.
\newblock Mint: Evaluating llms in multi-turn interaction with tools and language feedback.
\newblock In \emph{The Twelfth International Conference on Learning Representations}.

\bibitem[{Xiao et~al.(2024)Xiao, Li, Xie, Getzen, Fang, Long, and Su}]{xiao2024algorithmic}
Jiancong Xiao, Ziniu Li, Xingyu Xie, Emily Getzen, Cong Fang, Qi~Long, and Weijie~J. Su. 2024.
\newblock \href {https://arxiv.org/abs/2405.16455} {On the algorithmic bias of aligning large language models with rlhf: Preference collapse and matching regularization}.
\newblock \emph{Preprint}, arXiv:2405.16455.

\bibitem[{Xu et~al.(2020)Xu, Zhong, Yepes, and Lau}]{xu2020forget}
Ying Xu, Xu~Zhong, Antonio Jose~Jimeno Yepes, and Jey~Han Lau. 2020.
\newblock Forget me not: Reducing catastrophic forgetting for domain adaptation in reading comprehension.
\newblock In \emph{2020 International Joint Conference on Neural Networks (IJCNN)}, pages 1--8. IEEE.

\bibitem[{Zhang et~al.(2023)Zhang, Press, Merrill, Liu, and Smith}]{zhang2023language}
Muru Zhang, Ofir Press, William Merrill, Alisa Liu, and Noah~A Smith. 2023.
\newblock How language model hallucinations can snowball.
\newblock \emph{arXiv preprint arXiv:2305.13534}.

\bibitem[{Zhou et~al.(2023)Zhou, Liu, Xu, Iyer, Sun, Mao, Ma, Efrat, Yu, YU, Zhang, Ghosh, Lewis, Zettlemoyer, and Levy}]{zhou2023lima}
Chunting Zhou, Pengfei Liu, Puxin Xu, Srini Iyer, Jiao Sun, Yuning Mao, Xuezhe Ma, Avia Efrat, Ping Yu, LILI YU, Susan Zhang, Gargi Ghosh, Mike Lewis, Luke Zettlemoyer, and Omer Levy. 2023.
\newblock \href {https://openreview.net/forum?id=KBMOKmX2he} {{LIMA}: Less is more for alignment}.
\newblock In \emph{Thirty-seventh Conference on Neural Information Processing Systems}.

\end{thebibliography}

\appendix

\section{Sankey Diagrams}
\label{sec:appendix-sankey}

\begin{figure*}[!ht]
\centering \footnotesize

\begin {minipage}{\textwidth} \centering
\includegraphics[width=1\linewidth]{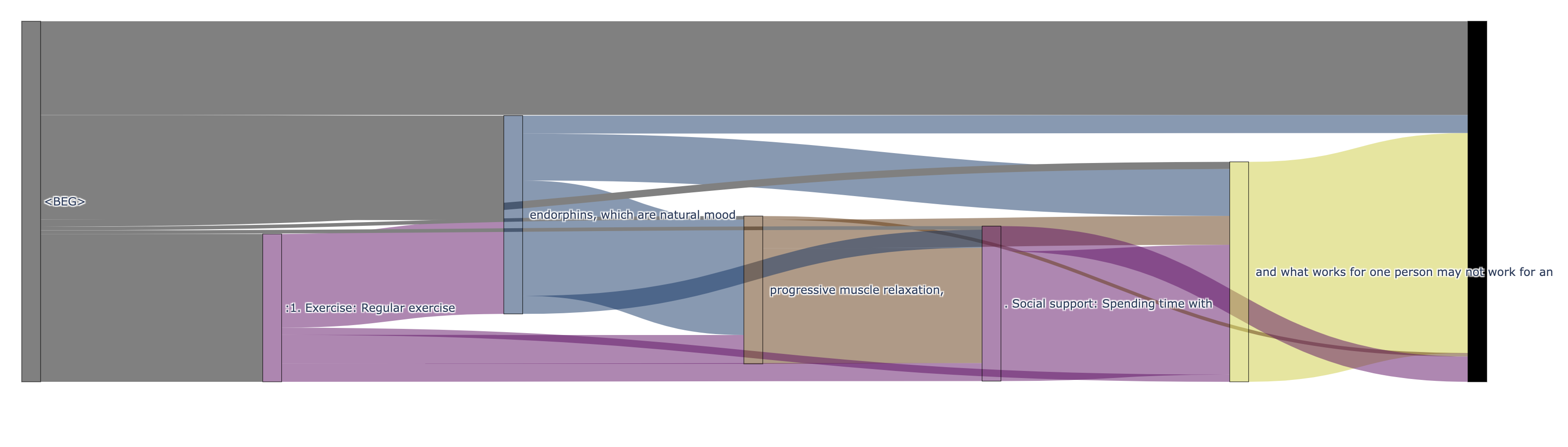}
 \textbf{Llama 2 70B RLHF, Nucleus Sampling, $p$=0.9}
\end{minipage}
\begin{minipage}{\textwidth} \centering
\includegraphics[width=1\linewidth]{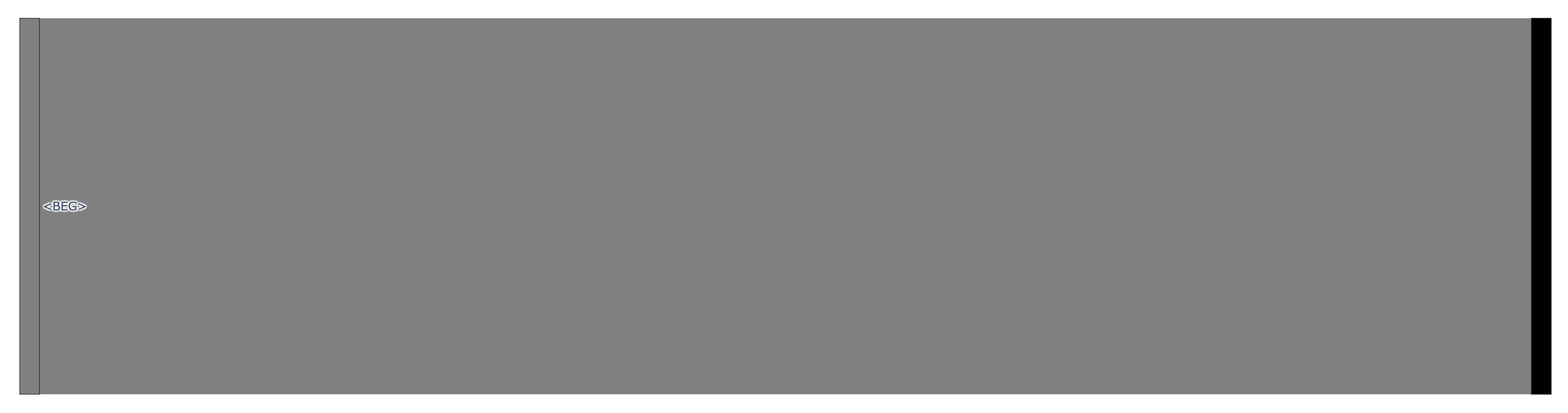}
 \textbf{Llama 2 70B Base, Nucleus Sampling, $p$=0.7}
\end{minipage}
\begin {minipage}{\textwidth} \centering
\includegraphics[width=1\linewidth]{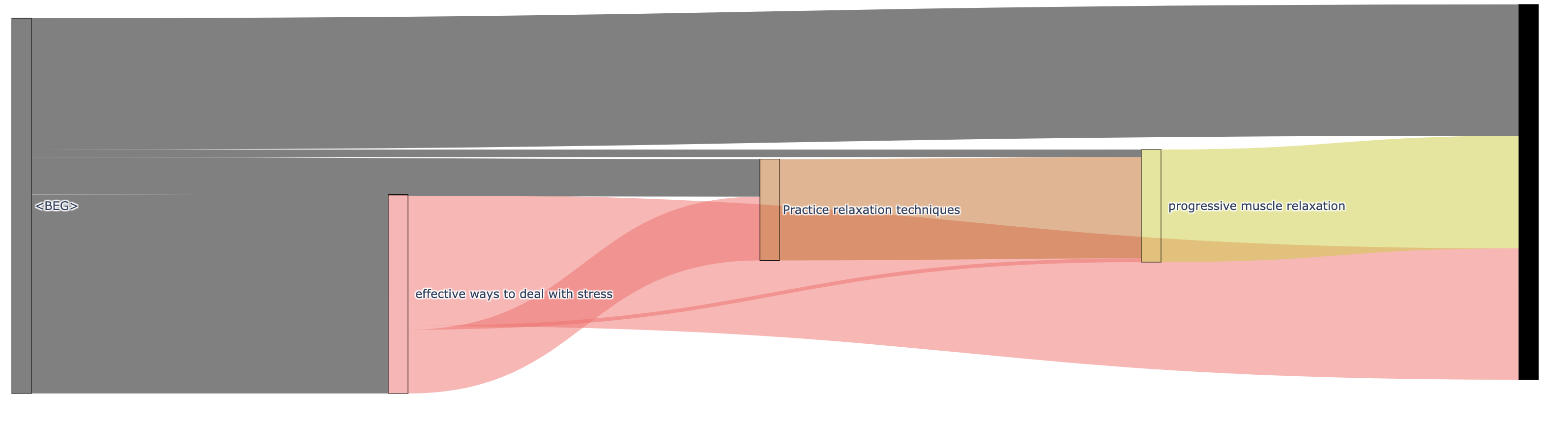}
 \textbf{Llama 2 7B RLHF, Nucleus Sampling, $p$=0.9}
\end{minipage}
\begin{minipage}{\textwidth} \centering
\includegraphics[width=1\linewidth]{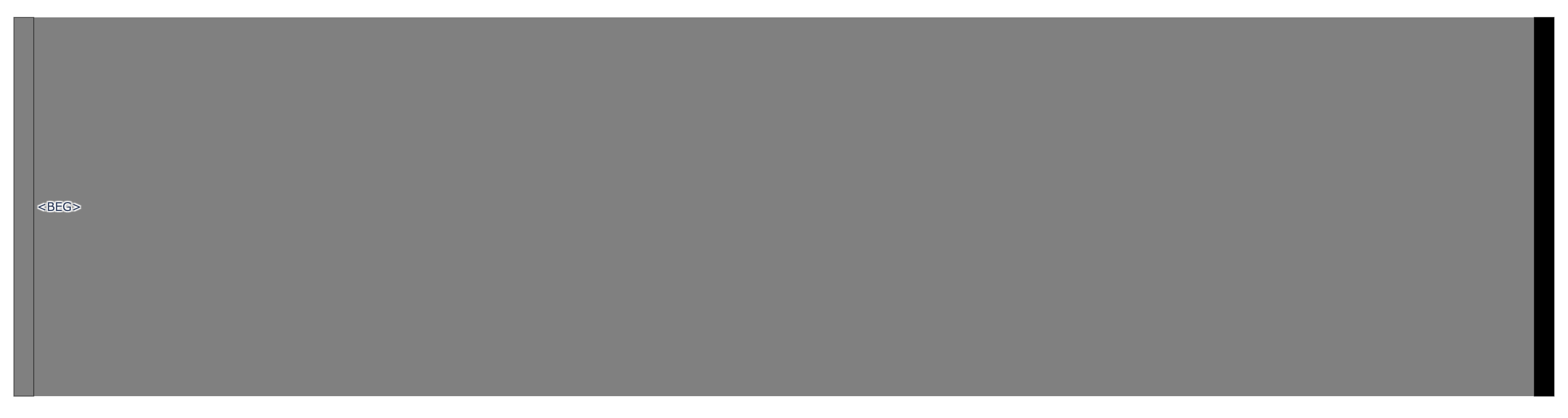}
 \textbf{Llama 2 7B Base, Nucleus Sampling, $p$=0.7}
\end{minipage}

\caption{
(\S\ref{sec:anchor_spans}) Further examples of Sankey diagrams, as in Figure~\ref{fig:sankey}. Each Sankey diagram visualizes 100 responses (nucleus sampling, p = 0.9 for RLHF and p = 0.7 for Base) to the prompt ``What are the most effective ways to deal with stress?''
}
\label{fig:sankey-app-1}
\end{figure*}

\begin{figure*}[t!h]
\centering \footnotesize

\begin {minipage}{\textwidth} \centering
\includegraphics[width=1\linewidth]{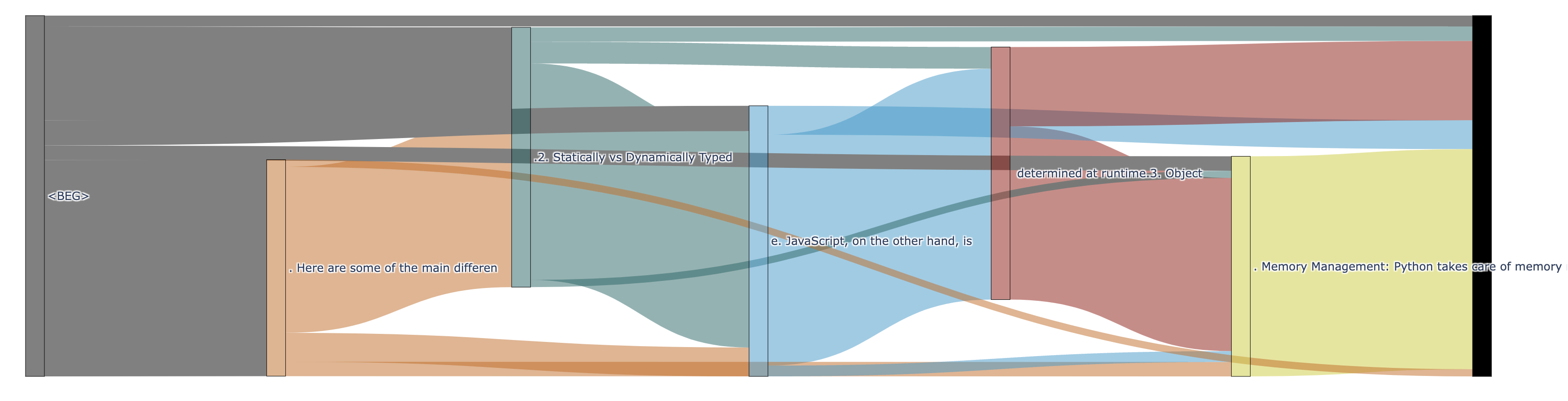}
 \textbf{Llama 2 70B RLHF, Nucleus Sampling, $p$=0.9}
\end{minipage}
\begin{minipage}{\textwidth} \centering
\includegraphics[width=1\linewidth]{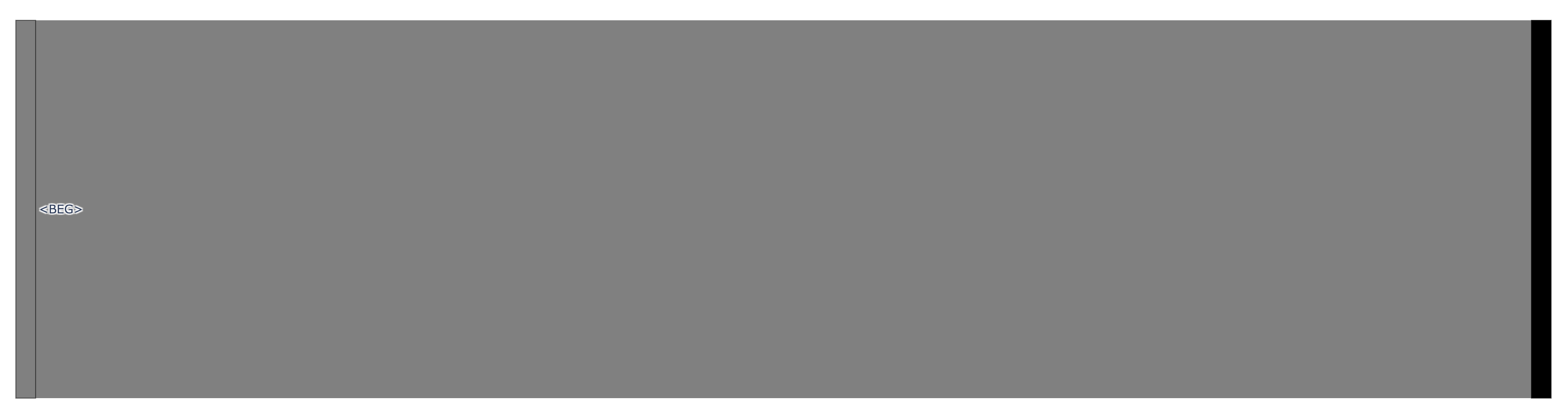}
 \textbf{Llama 2 70B Base, Nucleus Sampling, $p$=0.7}
\end{minipage}
\begin {minipage}{\textwidth} \centering
\includegraphics[width=1\linewidth]{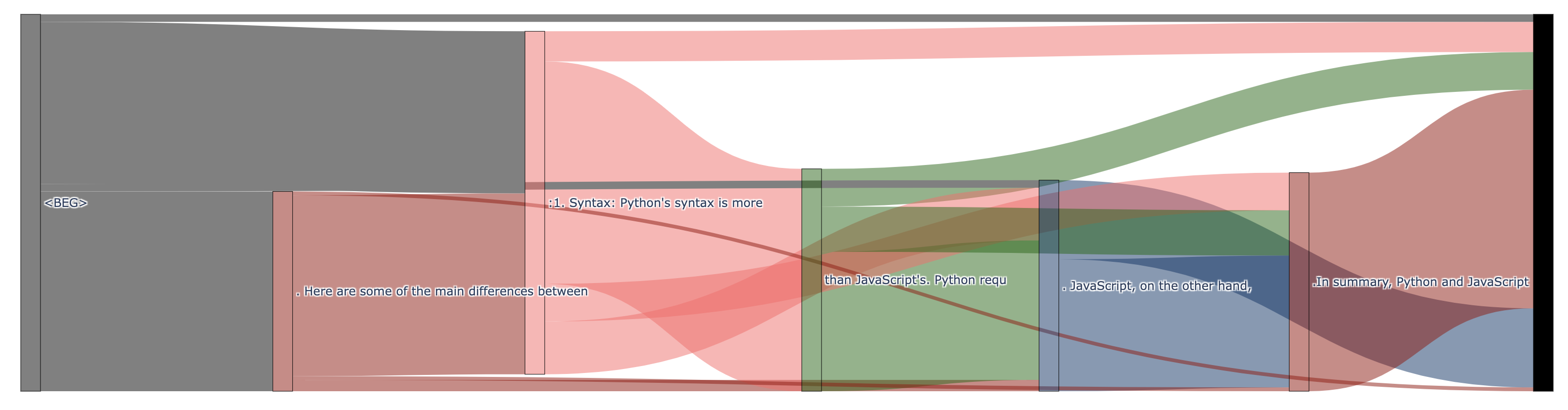}
 \textbf{Llama 2 7B RLHF, Nucleus Sampling, $p$=0.9}
\end{minipage}
\begin{minipage}{\textwidth} \centering
\includegraphics[width=1\linewidth]{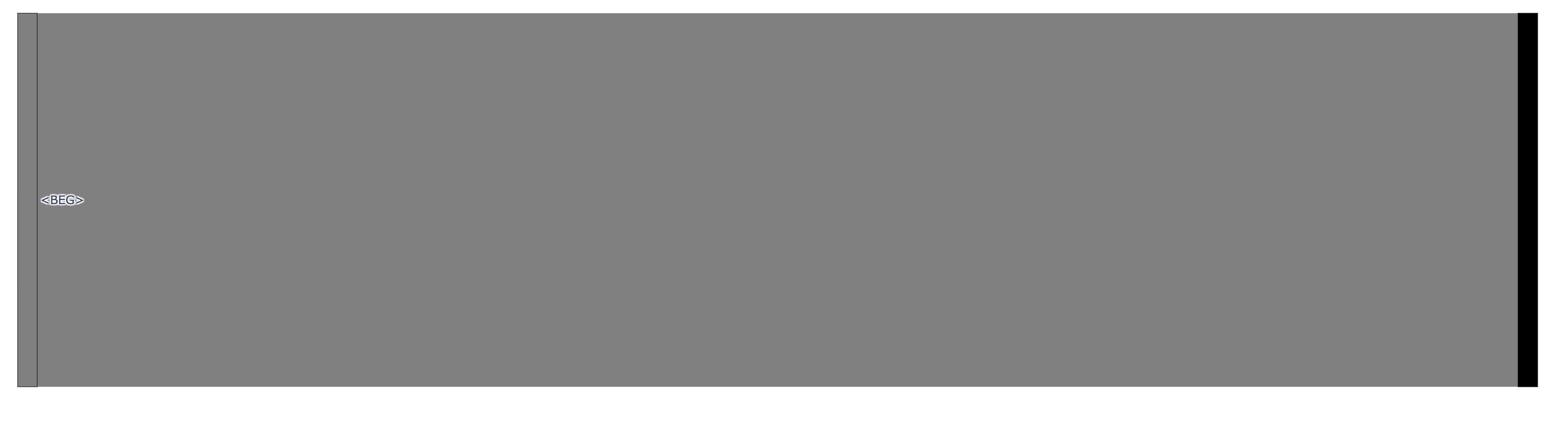}
 \textbf{Llama 2 7B Base, Nucleus Sampling, $p$=0.7}
\end{minipage}

\caption{
(\S\ref{sec:anchor_spans}) Further examples of Sankey diagrams, as in Figure~\ref{fig:sankey}. Each Sankey diagram visualizes 100 responses (nucleus sampling, p = 0.9 for RLHF and p = 0.7 for Base) to the prompt ``What are the main differences between Python and JavaScript programming languages?''
}
\label{fig:sankey-app-2}
\end{figure*}

\begin{figure*}[t!h]
\centering \footnotesize

\begin {minipage}{\textwidth} \centering
\includegraphics[width=1\linewidth]{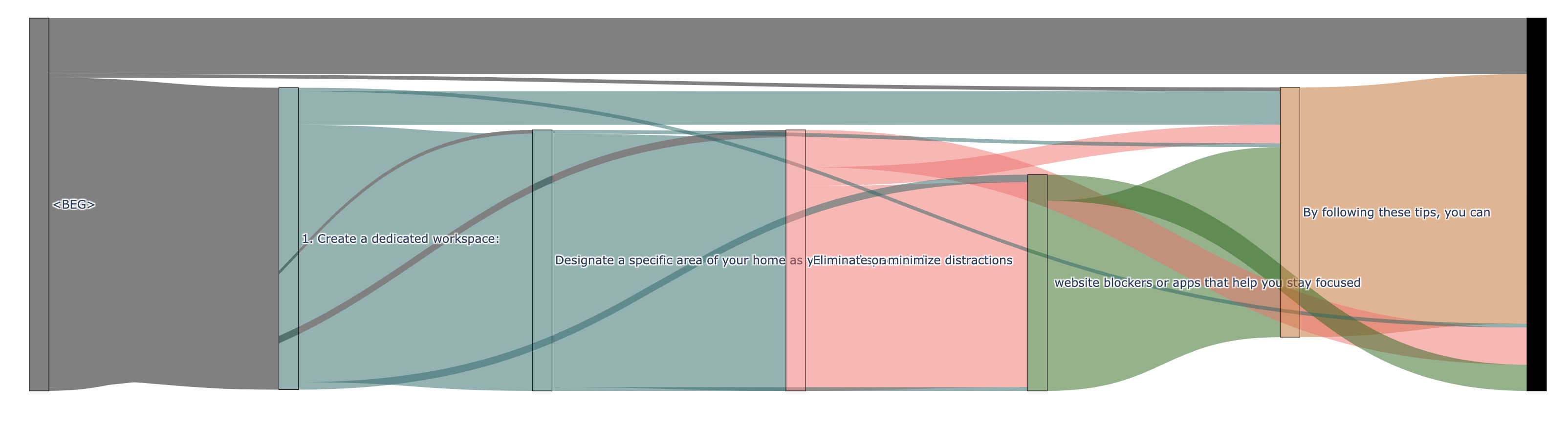}
 \textbf{Llama 2 70B RLHF, Nucleus Sampling, $p$=0.9}
\end{minipage}
\begin{minipage}{\textwidth} \centering
\includegraphics[width=1\linewidth]{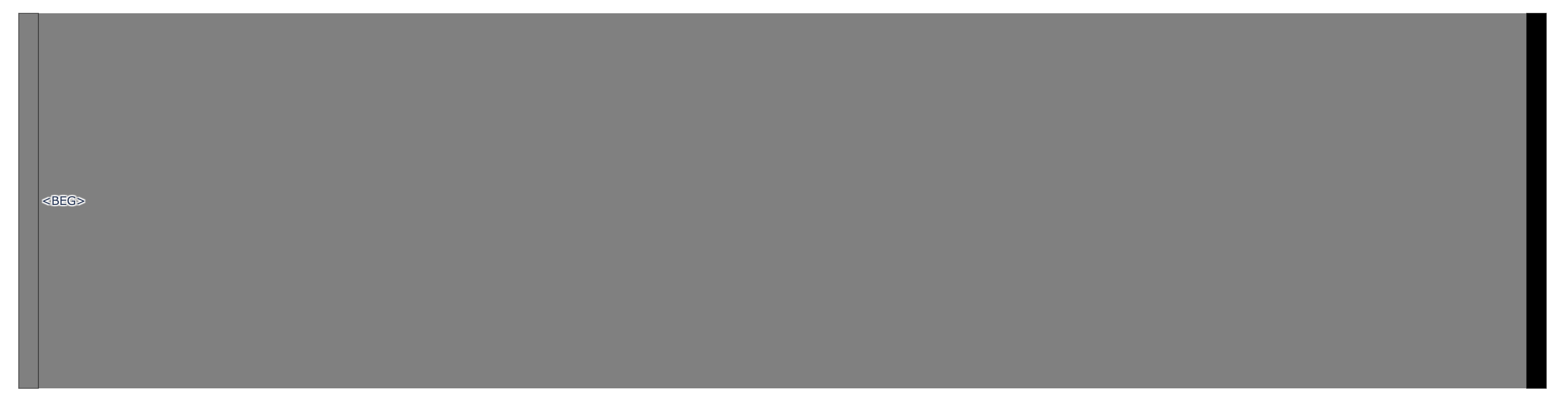}
 \textbf{Llama 2 70B Base, Nucleus Sampling, $p$=0.7}
\end{minipage}
\begin {minipage}{\textwidth} \centering
\includegraphics[width=1\linewidth]{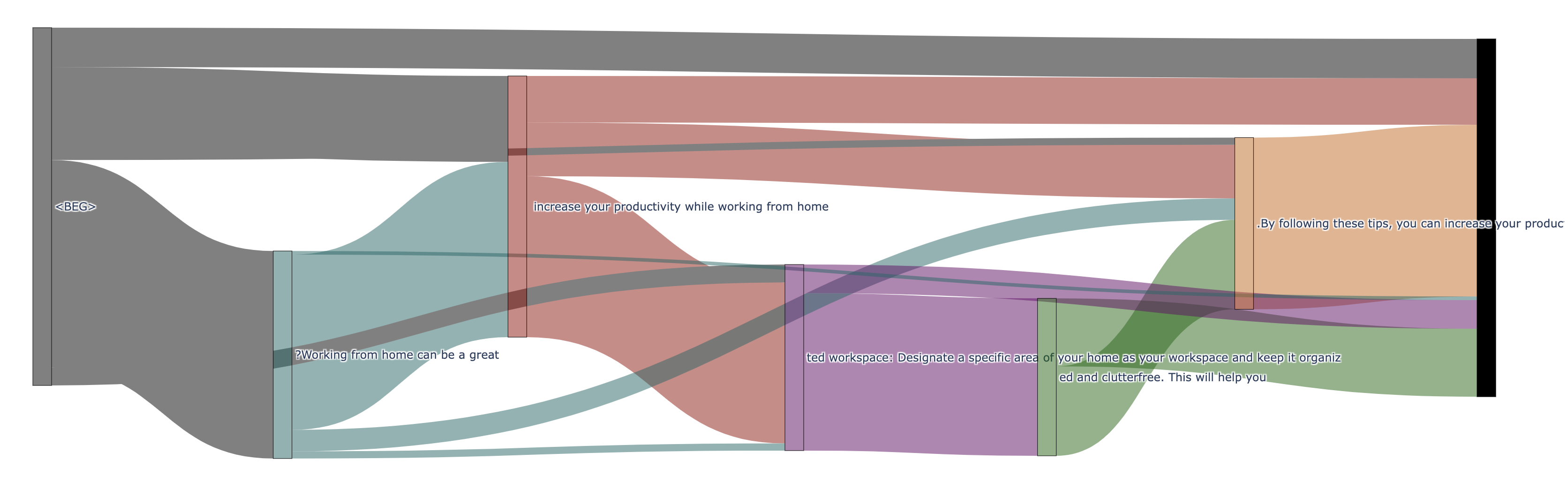}
 \textbf{Llama 2 7B RLHF, Nucleus Sampling, $p$=0.9}
\end{minipage}
\begin{minipage}{\textwidth} \centering
\includegraphics[width=1\linewidth]{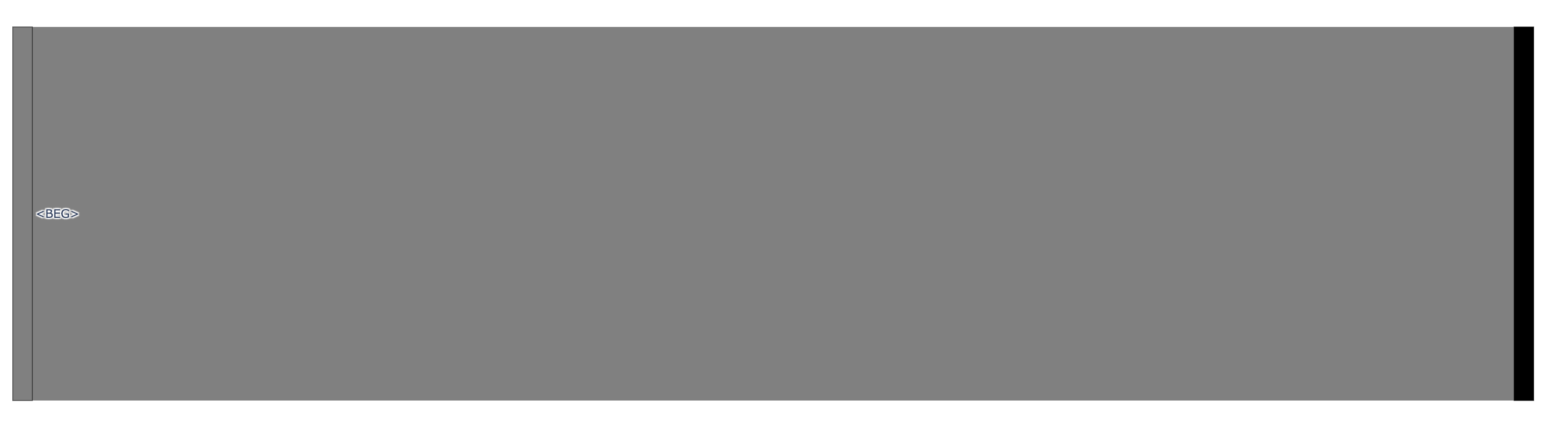}
 \textbf{Llama 2 7B Base, Nucleus Sampling, $p$=0.7}
\end{minipage}

\caption{
(\S\ref{sec:anchor_spans}) Further examples of Sankey diagrams, as in Figure~\ref{fig:sankey}. Each Sankey diagram visualizes 100 responses (nucleus sampling, p = 0.9 for RLHF and p = 0.7 for Base) to the prompt ``How can I increase my productivity while working from home?''
}
\label{fig:sankey-app-3}
\end{figure*}

\begin{figure*}[t!h]
\centering \footnotesize

\begin {minipage}{\textwidth} \centering
\includegraphics[width=1\linewidth]{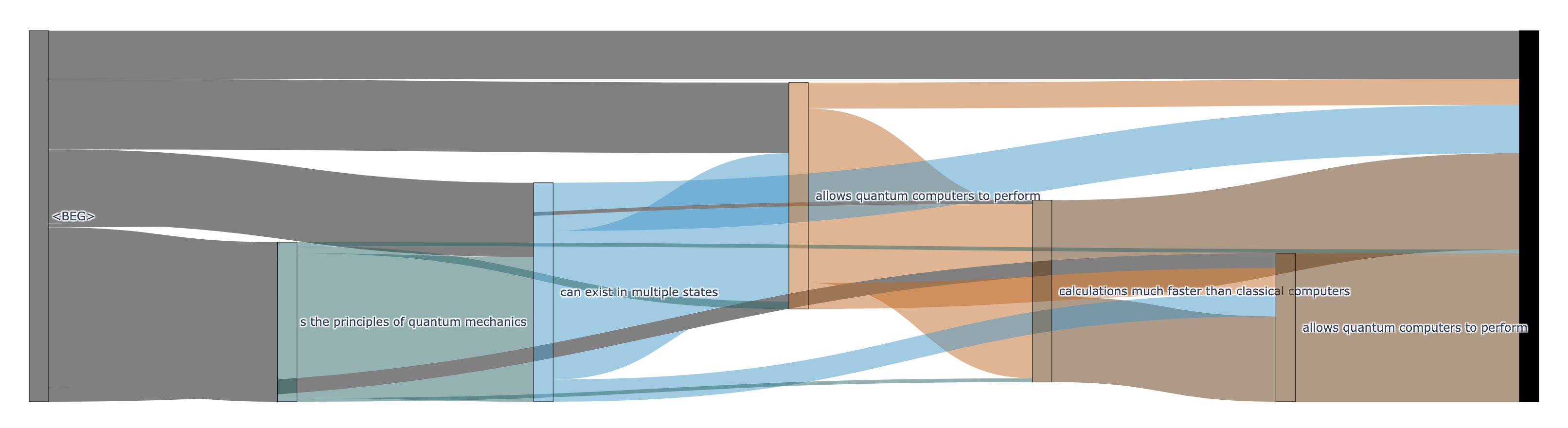}
 \textbf{Llama 2 70B RLHF, Nucleus Sampling, $p$=0.9}
\end{minipage}
\begin{minipage}{\textwidth} \centering
\includegraphics[width=1\linewidth]{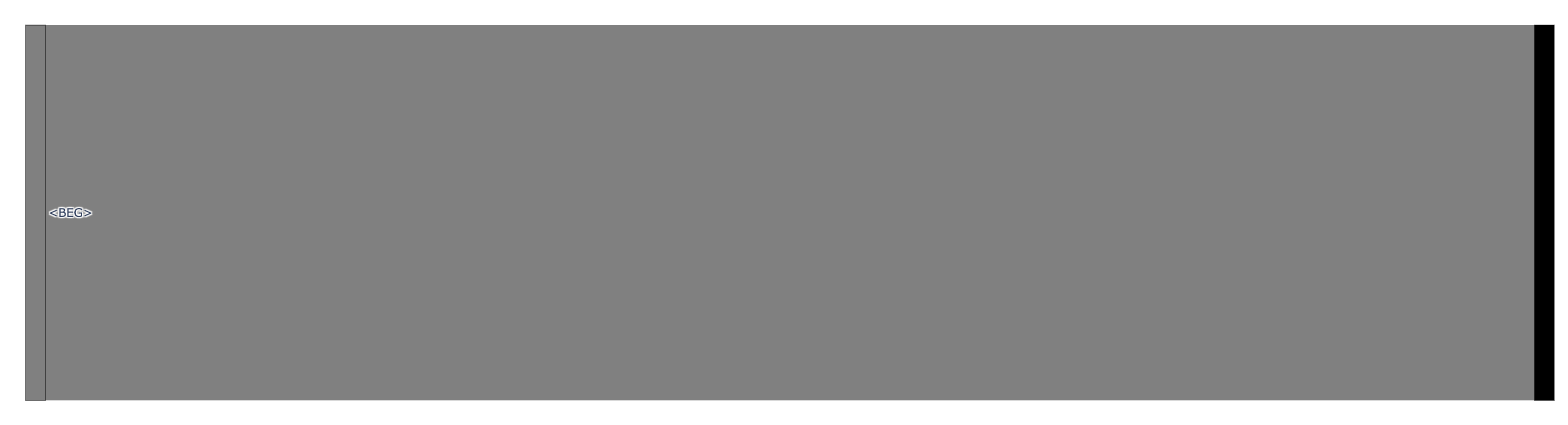}
 \textbf{Llama 2 70B Base, Nucleus Sampling, $p$=0.7}
\end{minipage}
\begin {minipage}{\textwidth} \centering
\includegraphics[width=1\linewidth]{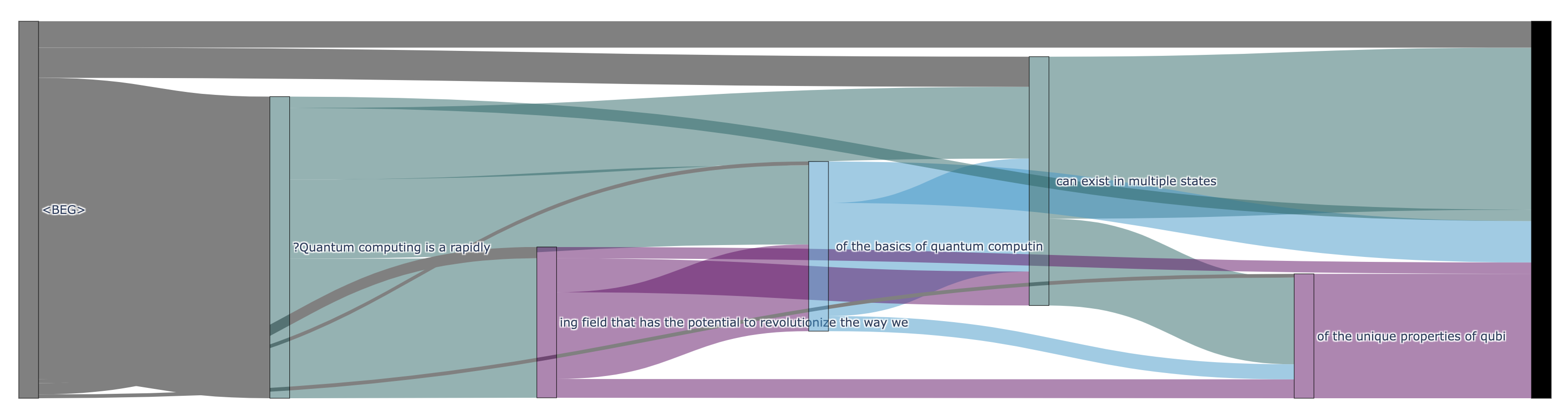}
 \textbf{Llama 2 7B RLHF, Nucleus Sampling, $p$=0.9}
\end{minipage}
\begin{minipage}{\textwidth} \centering
\includegraphics[width=1\linewidth]{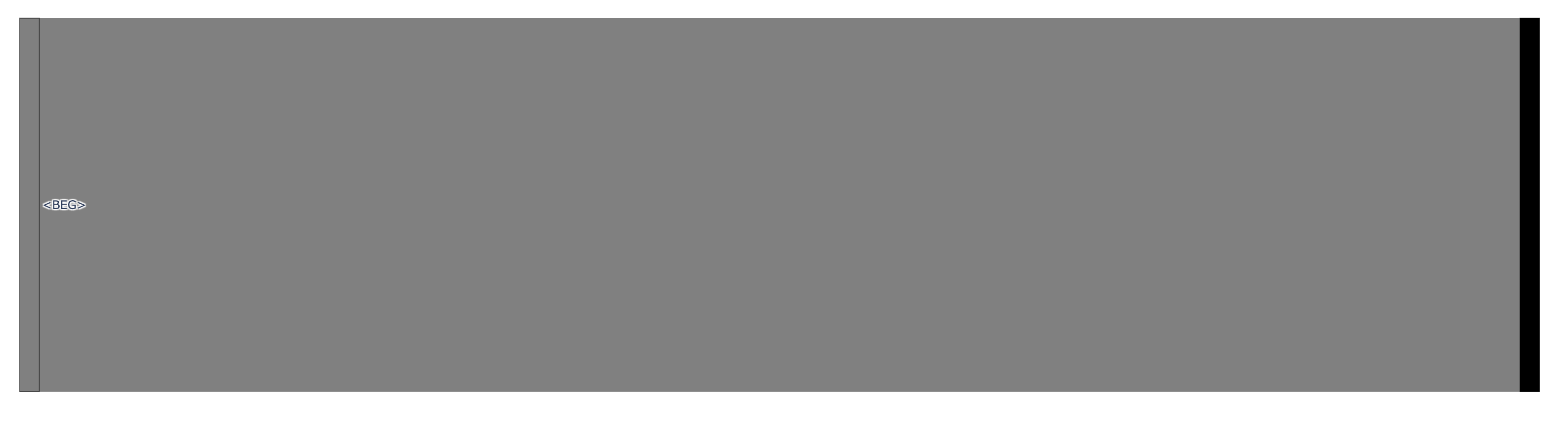}
 \textbf{Llama 2 7B Base, Nucleus Sampling, $p$=0.7}
\end{minipage}

\caption{
(\S\ref{sec:anchor_spans}) Further examples of Sankey diagrams, as in Figure~\ref{fig:sankey}. Each Sankey diagram visualizes 100 responses (nucleus sampling, p = 0.9 for RLHF and p = 0.7 for Base) to the prompt ``Can you explain the basics of quantum computing?''
}
\label{fig:sankey-app-4}
\end{figure*}

Figures \ref{fig:sankey-app-1}, \ref{fig:sankey-app-2}, \ref{fig:sankey-app-3}, \ref{fig:sankey-app-4} show further Sankey Diagrams—sampled at random from the 80 Vicuna prompts.


\clearpage

\section{Overlap Graphs}
\label{app:overlap}

\begin{figure*}[b!h]
\centering \footnotesize

\begin {minipage}{\textwidth} \centering
\includegraphics[width=1\linewidth]{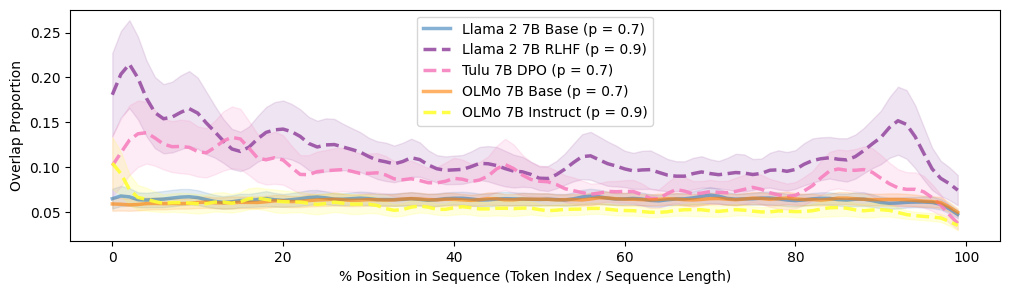}
\end{minipage}
\begin{minipage}{\textwidth} \centering
\includegraphics[width=1\linewidth]{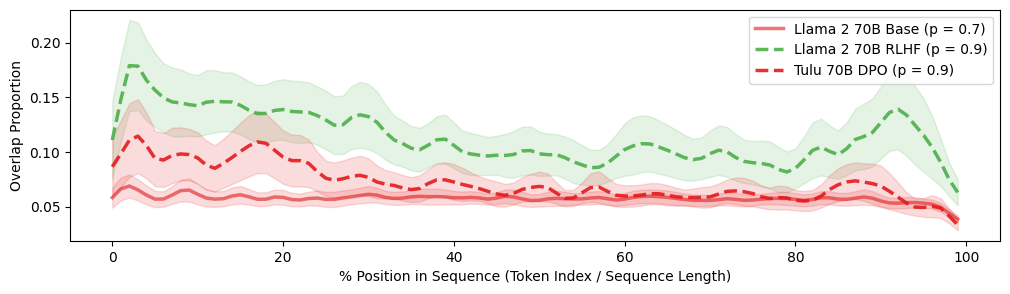}
\end{minipage}

\caption{
\textbf{RLHF model generations on the same prompt are highly similar to each other, unlike Base LMs.} For each of 80 short prompts, we collect and align 100 generations (nucleus sampling, p = 0.9) from Base (pretrained) and RLHF models. (\S\ref{sec:backbones}) Over the sequence length, the number of generations aligned with at least 5 others, averaged over all prompts. Base model generations maintain low levels of alignment. RLHF model generations exhibit high alignment throughout. \textbf{Above:} (\S\ref{sec:anchor_spans}) LLMs with 7B parameters \textbf{Below:} LLMs with 70B parameters.
}
\label{fig:overlap-2}
\end{figure*}

\begin{figure*}[b!h]
\centering \footnotesize

\begin {minipage}{\textwidth} \centering
\includegraphics[width=1\linewidth]{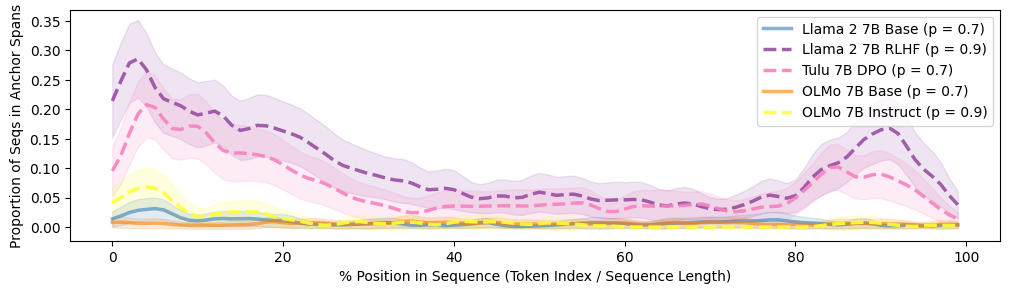}
\end{minipage}
\begin{minipage}{\textwidth} \centering
\includegraphics[width=1\linewidth]{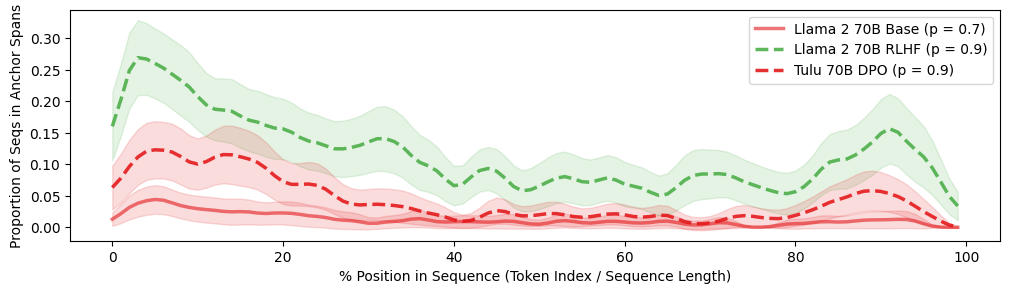}
\end{minipage}

\caption{
\textbf{RLHF model generations on the same prompt are highly similar to each other, unlike Base LMs.} For each of 80 short prompts, we collect and align 100 generations (nucleus sampling, p = 0.9) from Base (pretrained) and RLHF models. (\S\ref{sec:backbones}) While Figures~\ref{fig:fig1}~and~\ref{fig:anchor-overlap} look at alignment on raw characters, these diagrams show that when considering directly what proportion of sequences are part of an \textit{anchor span} (see \S\ref{sec:anchor_spans}) the pattern remains. \textbf{Above:} (\S\ref{sec:anchor_spans}) LLMs with 7B parameters \textbf{Below:} LLMs with 70B parameters.
}
\label{fig:anchor-overlap}
\end{figure*}

Figure~\ref{fig:overlap-2} plots the same overlap metrics as Figure~\ref{fig:fig1}~(top), but for more models. Figure~\ref{fig:anchor-overlap} shows that these metrics show a similar story when measuring the proportion of sequences that participate in \textit{anchor spans} (see \S\ref{sec:anchor_spans}) rather than looking at string alignment directly.

\clearpage

\clearpage

\section{Ngram Charts}
\label{app:ngrams}

\begin{figure*}
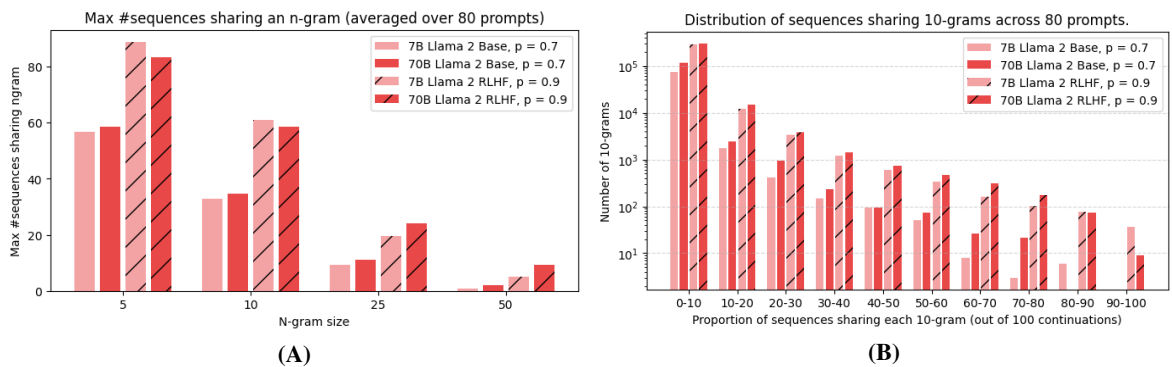

\centering \footnotesize

\begin{minipage}{0.48\textwidth}
 \centering
 \includegraphics[width=1\linewidth]{img/ngrams_a.png}
 \textbf{(A)}
\end{minipage}
\begin {minipage}{0.48\textwidth}
 \centering
 \includegraphics[width=1\linewidth]{img/ngrams_b.png}
 \textbf{(B)}
\end{minipage}

\caption{
\textbf{(\S\ref{sec:backbones}) Given a short prompt, RLHF models heavily reuse n-grams across many independently generated continuations (nucleus sampling, p = 0.9), with the most common 10-gram appearing in 60\% of generations on average.} For each of 80 short prompts, we collect 100 generations from Base and RLHF models using nucleus sampling with p = 0.7 and p = 0.9, respectively. \textbf{(A)} For each prompt, the number of generations, out of 100 total, which contain the most common n-gram ($n \in [5, 10, 25, 50]$), averaged across all prompts. \textbf{(B)} A histogram, binning 10-grams by the number of sequences containing that 10-gram. Counts are log-scale. Compared to Base models, RLHF models much more frequently generate the same 10-gram in nearly all continuations for a prompt. 
}
\label{fig:ngrams-2}
\end{figure*}

Figure~\ref{fig:ngrams-2} shows the same statistics as Figure~\ref{fig:ngrams} for further models.

\clearpage

\section{Shannon Game}
\label{app:shannon-game}

What if the ranking information about which tokens are more or less likely is still preserved in RLHF models, even if the probabilities themselves are distorted? 
The perplexities RLHF adapted LLMs yield are higher, but this could be potentially be a result of the distribution collapse RLHF models undergo (see \S\ref{sec:collapse}). To evaluate how well LLMs rank the gold next token, we evaluate both Base and RLHF models using the Shannon Game. 

\paragraph{Experimental setup} The Shannon Game \cite{hovy1998automated}, is a next token prediction task, except no probabilities are used. Instead, models are judged by the number of \textit{incorrect guesses} that a model ranks with a higher \textit{score} over the target token. The Shannon Game is invariant to relative differences in exactly how much probability is allocated to different strings, and is only sensitive to the \textit{ordering} that tokens are given in the hypothesis.  We evaluate the Base Llama 2 and RLHF Llama 2 (the Chat version) on LAMBADA~\citep{paperno-EtAl:2016:P16-1}, a collection of narrative passages designed to test the ability of LLMs on predicting the final word of a whole passage.

\paragraph{Results} Table~\ref{table-shannon} shows that this is not the case via the Shannon Game, revealing that RLHF adapted LLMs are worse at ranking possible next-tokens, not just assigning them probability. Table~\ref{table-shannon} shows that RLHF models are worse at the Shannon Game, suggesting that the ability to model arbitrary aspects of the textual world are diminished by current agent adaptation techniques. 

While it is tempting to assume that this is merely a result of imperfect RLHF methods, we argue that this trade-off is inherent to agent-adaptation. To generate multiple hundreds of tokens towards a singular goal, an \agent{} must limit the amount of \textit{uncertainty about future tokens}. Planning a coherent document while marginalizing over all possible paths is an \textit{exponentially} harder problem then collapsing onto a small set of possibilities. We hypothesize that the long-form generation capabilities of current RLHF models, are a general feature \agents{}: limiting the subspace of possibilities for any given prompt allows for better planning within this subspace. Further evidence for this hypothesis is given in \S\ref{sec:planning}.

\begin{table}[t]
\begin{center}
\begin{tabular}{llrrr}
\toprule
Llama 2 & Condition &    EM &    F1 &  Avg. guesses \\
\midrule
    \multirow{2}{*}{7B} &      Base & 68.41 & 70.31 &          7.62 \\
    &      RLHF & 56.70 & 60.78 &          9.99 \\
    \hdashline
    \multirow{2}{*}{70B} &      Base & 67.60 & 73.10 &          6.18 \\
   &      RLHF & 66.47 & 67.50 &          8.64 \\
\bottomrule
\end{tabular}
\end{center}
\caption{\textbf{RLHF models are worse than Base models at ranking possible next tokens on the LAMBADA dataset}~\citep{paperno-EtAl:2016:P16-1}, requiring more incorrect guesses to identify the correct token (Avg. guesses). \label{table-shannon}}
\end{table}

\section{Experimental Setups}
\label{sec:setup-details}

\subsection{Perplexity Datasets}
We list the corpora used in our perplexity experiments (\S\ref{sec:notllms}) in \autoref{tbl: ppl_data}. 

\begin{table}[h]
\centering
\small
\caption{Data used for perplexity experiments in \S\ref{sec:notllms} } \label{tbl: ppl_data}
\begin{tabular}{@{}ll@{}}
\toprule
Category            & Data
\\ \midrule
\multirow{3}{*}{Pretraining}         & Wikipedia                                           \\
                    & C4 \citep{roberts2019exploring}                     \\
                    & Arxiv \citep{clement2019arxiv}                      \\ \midrule
\multirow{2}{*}{New Corpus}            & New BBC \citep{li2023avoiding}                      \\
                    & New Arxiv \citep{li2023avoiding}                    \\ \midrule
Instruction Data   & Humpback \citep{li2023self}                         \\ 

\midrule
\multirow{3}{*}{Chat}                 & Anthropic Harmless \citep{bai2022training}          \\
                    & Anthropic Helpful \citep{bai2022training}          \\
                    & OASST1 \citep{kopf2024openassistant}               \\ \bottomrule
\end{tabular}
\end{table} 

\subsection{Self-Perplexity Datasets}
We list the corpora used in our self-perplexity experiments (\S\ref{sec:collapse}) in \autoref{tbl: self_ppl_data}.
\begin{table}[h]
\small
\centering
\caption{Data used for self-perplexity experiments in \S\ref{sec:collapse} } \label{tbl: self_ppl_data}
\begin{tabular}{@{}ll@{}}
\toprule
Category            & Data
\\ \midrule
\multirow{3}{*}{Pretraining}         & Wikipedia                                           \\
                    & C4 \citep{roberts2019exploring}                     \\
                    & Arxiv \citep{clement2019arxiv}                      \\ \midrule
\multirow{2}{*}{New Corpus}            & New BBC \citep{li2023avoiding}                      \\
                    & New Arxiv \citep{li2023avoiding}                    \\ \midrule
\multirow{2}{*}{Instruction Data}    & Humpback \citep{li2023self}      \\
& LIMA \citep{zhou2023lima}
\\ \midrule
\multirow{4}{*}{Chat}                 & Anthropic Harmless \citep{bai2022training}          \\
                    & Anthropic Helpful \citep{bai2022training}          \\
                    & OASST1 \citep{kopf2024openassistant} \\
                 &   Vicuna \citep{vicuna2023}
                    \\ \bottomrule
\end{tabular}
\end{table}

\subsection{Setup for distribution collapse experiments} \label{sec:setup-topk}
To demonstrate agent model collapse, we checked cumulative probability of the most probable k tokens (Figure~\ref{fig:cum-probs}) and percentage of non-negligible vocabulary (Figure~\ref{fig:non-neg}) for RLHF and Base models on gold vs. their generations, averaged across the diverse dataset used in \autoref{tbl: self_ppl_data}. 



\begin{figure}[t]
\centering \footnotesize
\includegraphics[width=0.75\linewidth]{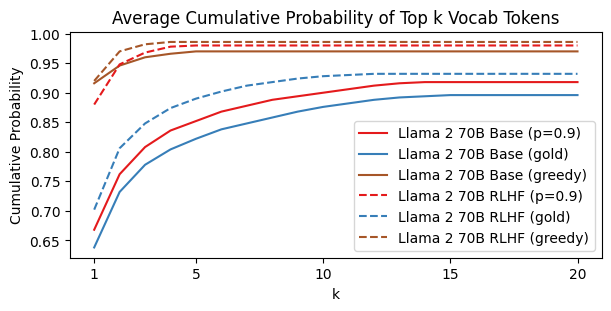}
\caption{
\textbf{\S\ref{sec:collapse} RLHF models assign nearly all of the next-token probability mass to a single token, more than Base models.} For Base and RLHF models, we calculate the next-token probability distributions on the gold sequences, as well as on the models' own generations (nucleus sampling, p=0.9; and greedy). We show the cumulative probability mass of the tokens, sorted in descending order of probability. RLHF models assign a larger portion of the probability mass to a very small number of tokens, compared to Base models.
}
\label{fig:cum-probs-with-greedy}
\end{figure}

\subsection{Planning experiments}
\label{sec:planning-experiments}
We use the Vicuna dataset \citep{vicuna2023} as a source of prompts and sample 100 continuations for each of the prompts from both 7B and 70B Base and RLHF models. 

\paragraph{Diversity of \textit{n}grams vs. nucleus sampling $p$.} We sample 100 continuations for each of the 80 prompts of the Vicuna dataset \citep{vicuna2023} and measure the proportion of unique \textit{n}grams (at the token level) for different  values of nucleus sampling $p$. We find that setting nucleus sampling $p=0.7$ for Base models achieves a similar diversity of \textit{n}grams as setting $p=0.9$ for RLHF models. We report results controlled for \textit{n}gram diversity by using these nucleus sampling values. Note that when using the same nucleus sampling values, the differences described are even more significant. 

\begin{table*}
\small
\resizebox{\textwidth}{!}{  
\begin{tabular}{@{}crrrrrrrrrr@{}}
\toprule
Model  & \multicolumn{5}{c}{Llama 2 70B}                                                                                                                                                       & \multicolumn{5}{c}{Llama 2 70B RLHF}                                                                                                                                                  \\ \midrule
ngrams & \multicolumn{1}{l}{p=0.6} & \multicolumn{1}{l}{\textbf{p=0.7}} & \multicolumn{1}{l}{p=0.8} & \multicolumn{1}{l}{p=0.9} & \multicolumn{1}{l}{p=1.0} & \multicolumn{1}{l}{p=0.6} & \multicolumn{1}{l}{p=0.7} & \multicolumn{1}{l}{p=0.8} & \multicolumn{1}{l}{\textbf{p=0.9}} & \multicolumn{1}{l}{p=1.0} \\ \midrule
1      & 4.11\%                           & \textbf{5.12\%}                           & 6.73\%                           & 9.20\%                           & 15.69\%                          & 3.35\%                           & 3.63\%                           & 3.89\%                           & \textbf{4.33\%}                           & 5.39\%                           \\
2      & 16.45\%                          & \textbf{21.34\%}                          & 29.34\%                          & 41.04\%                          & 60.61\%                          & 13.18\%                          & 14.70\%                          & 16.00\%                          & \textbf{17.96\%}                          & 22.59\%                          \\
3      & 29.10\%                          & \textbf{37.56\%}                          & 50.77\%                          & 68.09\%                          & 86.80\%                          & 23.67\%                          & 26.73\%                          & 29.24\%                          & \textbf{32.71\%}                          & 40.36\%                          \\
4      & 38.33\%                          & \textbf{48.54\%}                          & 63.34\%                          & 80.78\%                          & 94.22\%                          & 32.03\%                          & 36.31\%                          & 39.77\%                          & \textbf{44.17\%}                          & 53.23\%                          \\ \midrule
Model  & \multicolumn{5}{c}{Llama 2 7B}                                                                                                                                                        & \multicolumn{5}{c}{Llama 2 7B RLHF}                                                                                                                                                   \\ \midrule
ngrams & \multicolumn{1}{l}{p=0.6} & \multicolumn{1}{l}{\textbf{p=0.7}} & \multicolumn{1}{l}{p=0.8} & \multicolumn{1}{l}{p=0.9} & \multicolumn{1}{l}{p=1.0} & \multicolumn{1}{l}{p=0.6} & \multicolumn{1}{l}{p=0.7} & \multicolumn{1}{l}{p=0.8} & \multicolumn{1}{l}{\textbf{p=0.9}} & \multicolumn{1}{l}{p=1.0} \\ \midrule
1      & 3.11\%                           & \textbf{4.15\%}                           & 6.08\%                           & 9.34\%                           & 17.17\%                          & 3.06\%                           & 3.29\%                           & 3.67\%                           & \textbf{4.03\%}                           & 5.42\%                           \\
2      & 12.18\%                          & \textbf{17.29\%}                          & 26.83\%                          & 42.26\%                          & 64.11\%                          & 12.51\%                          & 13.78\%                          & 15.67\%                          & \textbf{17.29\%}                          & 23.08\%                          \\
3      & 21.57\%                          & \textbf{30.66\%}                          & 47.06\%                          & 70.26\%                          & 89.50\%                          & 23.43\%                          & 26.03\%                          & 29.64\%                          & \textbf{32.65\%}                          & 42.14\%                          \\
4      & 28.50\%                          & \textbf{39.79\%}                          & 59.04\%                          & 82.79\%                          & 95.86\%                          & 32.62\%                          & 36.32\%                          & 41.21\%                          & \textbf{45.14\%}                          & 56.26\%                          \\ \bottomrule

\end{tabular}}
\caption{Proportion of unique \textit{n}grams as a function of nucleus sampling $p$. \label{tab:ngramdiversity}}
\end{table*}



\paragraph{Alignment setup.} We employ the sequence alignment software MAFFT (Multiple Alignment using Fast Fourier Transform) \citep{katoh2013mafft} to align the 100 continuations of each prompt of the Vicuna dataset. The MAFFT software has a ``text'' setting designed to align multiple sequences with arbitrary characters. This is a method used in bioinformatics to align three or more biological sequences (generally protein, DNA, or RNA). In simple terms, it's a way of lining up these sequences to identify regions of similarity. The problem of aligning multiple generations is not trivial and, like in biological sequences, needs to handle deletions, insertions, substitutions, and translocations of tokens in the sequence. Biological alignment software is designed to handle these operations. In fact, we find MAFFT to be effective at aligning generations which might deviate at points due to some phrases being worded differently or omitted still often converge back later.

The plot in \autoref{fig:fig1} was obtained by applying MAFFT to the 100 generations for each prompt. We removed any position in the aligned output that was not shared by at least 5 sequences. We then calculate overlap by measuring the proportion of pairwise matches across sequences for each position. Next we downsampled the resulting sequence of overlap scores, which have varying lengths, to 100-dimensional sequence averaging corresponding values. Finally, we averaged across all prompts and applied a 1D Gaussian filter to smooth the curves. The shaded areas indicate confidence intervals for each position.

\end{document}